\documentclass{article}
\usepackage{booktabs}       
\usepackage{amsfonts}       
\usepackage{microtype}      
\usepackage{xcolor}
\usepackage{url}
\usepackage{verbatim} 
\usepackage{graphicx}
\usepackage{caption} 
\usepackage{multirow}
\usepackage[linesnumbered,ruled]{algorithm2e}
\usepackage{xspace}
\usepackage{epsfig}
\usepackage{amsmath}
\usepackage{amsthm}
\usepackage{amssymb}
\usepackage{times}
\usepackage{xr}
\usepackage{bbm}
\usepackage{dsfont}
\usepackage{bm}
\usepackage{subcaption}
\usepackage{enumitem}
\usepackage{hyperref}       

\newcommand{\note}[1]{{{\textcolor{blue}{[note: #1]}}}}
\newcommand{\jure}[1]{{{\textcolor{red}{[Jure: #1]}}}}
\newcommand{\rex}[1]{{{\textcolor{magenta}{[Rex: #1]}}}}
\newcommand{\will}[1]{{{\textcolor{orange}{[Will: #1]}}}}

\newcommand{\jiaxuan}[1]{{{\textcolor{green}{[Jiaxuan: #1]}}}}

\newcommand{\xhdr}[1]{{\noindent\bfseries #1}.}
\newcommand{\todo}[1]{{\textcolor{red}{TODO: #1}}}

\newcommand{\CITE}{{\textcolor{red}{[CITE]}}}
\newcommand{\name}{GraphRNN\xspace}

\newcommand{\R}{\mathbb{R}}

\newcommand{\cut}[1]{}

\newtheorem{proposition}{Proposition}

\newtheorem{corollary}{Corollary}

\newtheorem{observation}{\textbf{Observation}}

\newcommand{\eg}{\emph{e.g.}}
\newcommand{\ie}{\emph{i.e.}}

\newcommand{\ba}{Barab\'asi-Albert }
\newcommand{\er}{Erd\H{o}s-R\'enyi}

\newcommand{\mmd}{\mathrm{MMD}}
\newcommand{\graphrnnmlp}{GraphRNN-S}
\newcommand{\graphrnnrnn}{GraphRNN}

\usepackage[accepted]{icml2018}

\icmltitlerunning{GraphRNN: Generating Realistic Graphs with Deep Auto-regressive Models}

\begin{document}

\twocolumn[
\icmltitle{GraphRNN: Generating Realistic Graphs with Deep Auto-regressive Models}



\icmlsetsymbol{equal}{*}

\begin{icmlauthorlist}
\icmlauthor{Jiaxuan You}{equal,st}
\icmlauthor{Rex Ying}{equal,st}
\icmlauthor{Xiang Ren}{usc}
\icmlauthor{William L. Hamilton}{st}
\icmlauthor{Jure Leskovec}{st}
\end{icmlauthorlist}

\icmlaffiliation{st}{Department of Computer Science, Stanford University, Stanford, CA, 94305}
\icmlaffiliation{usc}{Department of Computer Science, University of Southern California, Los
  Angeles, CA, 90007}

\icmlcorrespondingauthor{Jiaxuan You}{jiaxuan@stanford.edu}

\icmlkeywords{Generative Model, Machine Learning, ICML}

\vskip 0.3in
]

\printAffiliationsAndNotice{\icmlEqualContribution}

\begin{abstract}


Modeling and generating graphs is fundamental for studying networks in biology, engineering, and social sciences. However, modeling complex distributions over graphs and then efficiently sampling from these distributions is challenging due to the non-unique, high-dimensional nature of graphs and the complex, non-local dependencies that exist between edges in a given graph.
Here we propose GraphRNN, a deep autoregressive model that addresses the above challenges and approximates any distribution of graphs with minimal assumptions about their structure. 
GraphRNN learns to generate graphs by training on a representative set of graphs and decomposes the graph generation process into a sequence of node and edge formations, conditioned on the graph structure generated so far. 
In order to quantitatively evaluate the performance of GraphRNN, we introduce a benchmark suite of datasets, baselines and novel evaluation metrics based on Maximum Mean Discrepancy, which measure distances between sets of graphs.
Our experiments show that GraphRNN significantly outperforms all baselines, learning to generate diverse graphs that match the structural characteristics of a target set, while also scaling to graphs $50\times$ larger than previous deep models.

\cut{
Generative models are an important class of algorithms used to understand the intrinsic structure of data. Generating graph-structured data, however, has remained to be a challenging problem due to its non-unique vector representations and complex dependencies among nodes and edges.
We propose GraphRNN, an autoregressive model that approximates arbitrary distributions of graphs with minimal assumption on their structure. GraphRNN decomposes distribution of graphs into a sequence of distributions of nodes and edges conditioned on graph structure representations.We introduce random BFS ordering, resulting in drastic performance and efficiency gains. We show that unlike previous work, the model is capable of capturing arbitrary dependencies in graph, learning from multiple graphs, and runs efficiently on graphs 50$\times$ larger than data used in previous deep models. We comprehensively evaluate our model using proposed metrics based on Maximum Mean Discrepancy, and demonstrate significant improvement over state-of-the-art models.

\note{alternative}
}

\cut{
In many domains including bioinformatics and social network analysis, it is often important to capture the distribution of a set of observed graphs, and to sample from this distribution, using generative models. 
However, the task of learning generative graph models directly from data remains challenging due to the lack of a canonical way to map graphs to unique vector representations and the non-local dependencies that exist among nodes and edges.
In this work, we propose GraphRNN, an autoregressive model that approximates any distribution of graphs with minimal assumptions on their structure. 
GraphRNN decomposes the generation process of graphs into a sequence of additions of nodes and edges, conditioned on previously generated graph structure. 
We show that unlike previous work, GraphRNN captures arbitrary dependencies in graphs, and runs efficiently on graphs 100$\times$ larger than data used in previous deep models. 
We propose metrics based on Maximum Mean Discrepancy to evaluate similarity between sets of graphs. 
Comprehensive comparisons with state-of-the-art baselines  demonstrate that our approach can more efficiently learn to generate diverse graphs that exhibit key characteristics learned from a training set.
}
\end{abstract}


\section{Introduction and Related Work}

Generative models for real-world graphs have important applications in many domains, including modeling physical and social interactions, discovering new chemical and molecular structures, and constructing knowledge graphs.
Development of generative graph models has a rich history, and many methods have been proposed that can generate graphs based on a priori structural assumptions~\cite{newman2010networks}. 
However, a key open challenge in this area is developing methods that can directly {\em learn} generative models from an observed set of graphs. 
Developing generative models that can learn directly from data is an important step towards improving the fidelity of generated graphs, and paves a way for new kinds of applications, such as discovering new graph structures and completing evolving graphs. 

In contrast, traditional generative models for graphs (\eg, \ba model, Kronecker graphs, exponential random graphs, and stochastic block models) \cite{erdos1959random,leskovec2010kronecker,albert2002statistical,airoldi2008mixed,leskovec2007graph,robins2007introduction} are hand-engineered to model a particular family of graphs, and thus do not have the capacity to directly learn the generative model from observed data. For example, the \ba model is carefully designed to capture the scale-free nature of empirical degree distributions, but fails to capture many other aspects of real-world graphs, such as community structure.

Recent advances in deep generative models, such as variational autoencoders (VAE) \cite{kingma2013auto} and generative adversarial networks (GAN) \cite{goodfellow2014generative}, have made important progress towards generative modeling for complex domains, such as image and text data. 
Building on these approaches a number of deep learning models for generating graphs have been proposed~\cite{kipf2016variational,grovergraphite,simonovsky2018graphvae,li2018learning}.
For example, \citealt{simonovsky2018graphvae} propose a VAE-based approach, while \citealt{li2018learning} propose a framework based upon graph neural networks. 
However, these recently proposed deep models are either limited to learning from a single graph \cite{kipf2016variational,grovergraphite} or generating small graphs with 40 or fewer nodes \cite{li2018learning,simonovsky2018graphvae}---limitations that stem from three fundamental challenges in the graph generation problem:
\begin{itemize}[itemsep=0pt,topsep=0pt,leftmargin=11pt]
\item {\bf Large and variable output spaces:}
To generate a graph with $n$ nodes the generative model has to output $n^2$ values to fully specify its structure. Also, the number of nodes $n$ and edges $m$ varies between different graphs and a generative model needs to accommodate such complexity and variability in the output space.
\item {\bf Non-unique representations:}
In the general graph generation problem studied here, we want distributions over possible graph structures without assuming a fixed set of nodes (\eg, to generate candidate molecules of varying sizes). 
In this general setting, a graph with $n$ nodes can be represented by up to $n!$ equivalent adjacency matrices, each corresponding to a different, arbitrary node ordering/numbering. Such high representation complexity is challenging to model and makes it expensive to compute and then optimize objective functions, like reconstruction error, during training. 
For example, GraphVAE \cite{simonovsky2018graphvae} uses approximate graph matching to address this issue, requiring $O(n^4)$ operations in the worst case \cite{cho2014finding}.
\item {\bf Complex dependencies:}
Edge formation in graphs involves complex structural dependencies. For example, in many real-world graphs two nodes are more likely to be connected if they share common neighbors \cite{newman2010networks}. Therefore, edges cannot be modeled as a sequence of independent events, but rather need to be generated jointly, where each next edge depends on the previously generated edges. 
\citealt{li2018learning} address this problem using graph neural networks to perform a form of ``message passing''; however, while expressive, this approach takes $O(mn^2\mathrm{diam}(G))$ operations to generate a graph with $m$ edges, $n$ nodes and diameter $\mathrm{diam}(G)$.
\end{itemize}


\cut{
Among some recent attempts to resolve these challenges, GraphVAE \cite{simonovsky2018graphvae} attempts to overcome the challenge non-unique representations by inexact graph
matching \cite{cho2014finding} but the proposed algorithm has $O(|V|^4)$ time complexity, making it difficult to scale to real-world graphs.
In addition, GraphVAE requires the maximum size of graph to be pre-specified.
Inspired by graph
convolutional neural networks (GCN) \cite{kipf2016semi}, an alternative model proposed in \citealt{li2018learning} incrementally generates a graph using GCN-based operations; however, this approach results in a time complexity of $O(|V|^2|E| \mathrm{diam}(G))$ to generate a graph with $|E|$ edges, $|V|$ nodes and diameter $\mathrm{diam}(G)$.
Thus, despite the significant advances made by these recent algorithms, we still lack an approach that can learn to generate graphs with more than ${\sim}40$ nodes.
} 

\vspace{7pt}
\xhdr{Present work} 
Here we address the above challenges and present \textit{Graph Recurrent Neural Networks} (\textbf{\name}), a scalable framework for learning generative models of graphs. 
\name\ models a graph in an autoregressive (or recurrent) manner---as a sequence of additions of new nodes and edges---to capture the complex joint probability of all nodes and edges in the graph.
In particular, \name\ can be viewed as a hierarchical model, where a {\em graph-level RNN} maintains the state of the graph and generates new nodes, while an {\em edge-level RNN} generates the edges for each newly generated node. 
Due to its autoregressive structure, \name\ can naturally accommodate variable-sized graphs, and we introduce a breadth-first-search (BFS) node-ordering scheme to drastically improve scalability. 
This BFS approach alleviates the fact that graphs have non-unique representations---by collapsing distinct representations to unique BFS trees---and the tree-structure induced by BFS allows us to limit the number of edge predictions made for each node during training.
Our approach requires $O(n^2)$ operations on worst-case (\ie, complete) graphs, but we prove that our BFS ordering scheme permits sub-quadratic complexity in many cases.  
\cut{
reduces the number of possible sequences we need to consider by collapsing multiple node orderings to a unique BFS tree.
The tree-structure induced by BFS also allows us to limit the dimensionality of the $S^\pi_i$ vectors. }


\cut{
 (Section \ref{sec:proposed}).
Illustrative case studies  \ref{sec:case_study}) and comprehensive empirical analyses (Section \ref{sec:experiments}), demonstrate how expressive recurrent neural networks allow \name\ to account for complex, non-local dependencies between edges in a given graph.
}

In addition to the novel \name\ framework, we also introduce a comprehensive suite of benchmark tasks and baselines for the graph generation problem, with all code made publicly available\footnote{The code is available in \url{https://github.com/snap-stanford/GraphRNN}, the appendix is available in \url{https://arxiv.org/abs/1802.08773}.}.
A key challenge for the graph generation problem is quantitative evaluation of the quality of generated graphs.
Whereas prior studies have mainly relied on visual inspection or first-order moment statistics for evaluation, we provide a comprehensive evaluation setup by comparing graph statistics such as the degree distribution, clustering coefficient distribution and motif counts for two sets of graphs based on variants of the Maximum Mean Discrepancy (MMD) \cite{gretton2012kernel}.
This quantitative evaluation approach can compare higher order moments of graph-statistic distributions and provides a more rigorous evaluation than simply comparing mean values.

Extensive experiments on synthetic and real-world graphs of varying size demonstrate the significant improvement \name\ provides over baseline approaches, including the most recent deep graph generative models as well as traditional models. 
Compared to traditional baselines (\eg, stochastic block models), \name\ is able to generate high-quality graphs on all benchmark datasets, while the traditional models are only able to achieve good performance on specific datasets that exhibit special structures. 
Compared to other state-of-the-art deep graph generative models, \name\ is able to achieve superior quantitative performance---in terms of the MMD distance between the generated and test set graphs---while also scaling to graphs that are $50\times$ larger than what these previous approaches can handle.
Overall, \name\ reduces MMD by $80\%\text{-}90\%$ over the baselines on average across all datasets and effectively generalizes, achieving comparatively high log-likelihood scores on held-out data. 

 \cut{
Graph-structured data have a variety of key properties in different application domains. A set of graphs could be
characterized by power law degree distribution, specific network motifs, clustering coefficient, or
a combination of them \CITE. 
We test each of these meature to ensure that the model is flexible enough to fit all of these characteristics of graphs.
In addition to visual examination and domain specific measurements, 
we designed a general quantitative framework to comprehensively evaluate the quality of generated graph samples. 
We compare the graph statistics such as degree distribution, clustering coefficient
distribution and motif counts for two \emph{sets} of graphs based on variants of the Maximum Mean
Discrepancy (MMD) \CITE. 
This method can compare higher order of moments of graph statistics distribution instead of
average graph statistics which only corresponds to the first order moments.

Common ways of evaluating graph generative models include qualitative visual inspection of graph embeddings and statistics (eg. degree distribution), and domain specific evaluation. However, quantitative evaluation of graph generative models that is agnostic to downstream tasks are important in order to compare the performance of the models from a variety of domains. Existing quantitative evaluation often only uses first-order moment graph statistics (eg. distribution of average degree for each graph in the generated samples), which cannot capture the complexity of degree characteristics for each graph.
}


\section{Proposed Approach}
\label{sec:proposed}

We first describe the background and notation for building generative models of graphs, and then describe our autoregressive framework, \name.

\subsection{Notations and Problem Definition}

An undirected graph\footnote{We focus on undirected graphs. Extensions to directed graphs and graphs with features are discussed in the Appendix.} $G=(V, E)$ is defined by its node set $V=\{v_1,...,v_n\}$ and edge set $E=\{(v_i,v_j)|v_i,v_j\in V\}$. One common way to represent a graph is using an adjacency matrix, which requires a node ordering $\pi$ that maps nodes to rows/columns of the adjacency matrix. More precisely, $\pi$ is a permutation function over $V$ (\emph{i.e.}, $(\pi(v_1),...,\pi(v_n))$ is a permutation of $(v_1,...,v_n)$). 
We define $\Pi$ as the set of all $n!$ possible node permutations. Under a node ordering $\pi$, a graph $G$ can then be represented by the adjacency matrix $A^{\pi}\in\mathbb{R}^{n\times n}$, where $A^\pi_{i,j} = \mathds{1}[(\pi(v_i),\pi(v_j))\in E]$. Note that elements in the set of adjacency matrices $A^\Pi=\{A^\pi|\pi\in\Pi\}$ all correspond to the same underlying graph.

The goal of {\em learning generative models of graphs} is to learn a distribution $p_{model}(G)$ over graphs, based on a set of observed graphs $\mathbb{G}=\{G_1,...,G_s\}$ sampled from data distribution $p(G)$, where each graph $G_i$ may have a different number of nodes and edges. 
 When representing $G\in\mathbb{G}$, we further assume that we may observe any node ordering $\pi$ with equal probability, \ie,
$p(\pi)=\frac{1}{n!}, \forall\pi\in\Pi$.
Thus, the generative model needs to be capable of generating graphs where each graph could have exponentially many representations, which is distinct from previous generative models for images, text, and time series.

Finally, note that traditional graph generative models (surveyed in the introduction) usually assume a single input training graph. Our approach is more general and can be applied to a single as well as multiple input training graphs.

\subsection{A Brief Survey of Possible Approaches}\label{sec:possible}
We start by surveying some general alternative approaches for modeling $p(G)$, in order to highlight the limitations of existing non-autoregressive approaches and motivate our proposed autoregressive architecture.

\xhdr{Vector-representation based models}
	One na\"ive approach would be to represent $G$ by flattening $A^\pi$ into a vector in $\mathbb{R}^{n^2}$, which is then used as input to any off-the-shelf generative model, such as a VAE or GAN. 
   However, this approach suffers from serious drawbacks: it cannot naturally generalize to graphs of varying size, and requires training on all possible node permutations or specifying a canonical permutation, both of which require $O(n!)$ time in general.
   \cut{
    However, unlike CNN which crucially makes use of spacial locality of pixels, this representation makes minimal use of the connectivity information in network architecture, thus requiring far more parameters to be learned. 
    In addition, VAE assumes conditional independence of each output dimension, conditioned on the latent representation \cite{kingma2013auto}, hence cannot capture more complex dependency between nodes and edges. 
    GAN suffers from mode collapse problem\cite{goodfellow2014generative}, and its discriminator would have to determine real and fake graphs up to any permutations of the vector representation. }
    
\cut{
	\xhdr{Random-walk based models}
	Random walks are a widely used tool to model structural properties of graphs, and the task of generating graphs can be simplified to generating set of random walks $R = (v_{r_1},...,v_{r_k})$ that obeys the distribution of random walks in a training graph. 
   However, like the na\"ive vector representation approach, this approach suffers from the issue that there are different representations $R^{\pi}$ under different permutations $\pi$, and learning the joint probability distribution $p(R^{\pi},\pi)$ is intractable in general.
    .}

	\xhdr{Node-embedding based models}
	There have been recent successes in encoding a graph's structural properties into node embeddings \cite{hamilton2017representation}, and one approach to graph generation could be to define a generative model that decodes edge probabilities based on pairwise relationships between learned node embeddings (as in \citealt{kipf2016variational}).
	However, this approach is only well-defined when given a fixed-set of nodes, limiting its utility for the general graph generation problem, and approaches based on this idea are limited to learning from a single input graph \cite{kipf2016variational,grovergraphite}. 

\cut{
	\xhdr{Sequence representation based models}
	One represent a graph $G$ as sequences corresponding to a $A^\pi$ under permutation $\pi(\cdot)$. Using Recurrent Neural Network (RNN), the model can take arbitrary length sequence as input. The complexity of learning $p(G)$ is also greatly reduced through decomposing $p(G)$ into a product of distributions and parameter weight sharing. A recent work propose to represent a graph as a $O(E)$ sequence with GCN based transition function, with time complexity $O(|E| |V|^2 \mathrm{diam}(G))$. Our model adopts this formulations with novel approaches that significantly improve performances and reduce the ime complexity. 
  }

\subsection{GraphRNN: Deep Generative Models for Graphs}\label{sec:autoregressive}

\begin{figure}[t]
  \centering
  \includegraphics[width=0.47\textwidth]{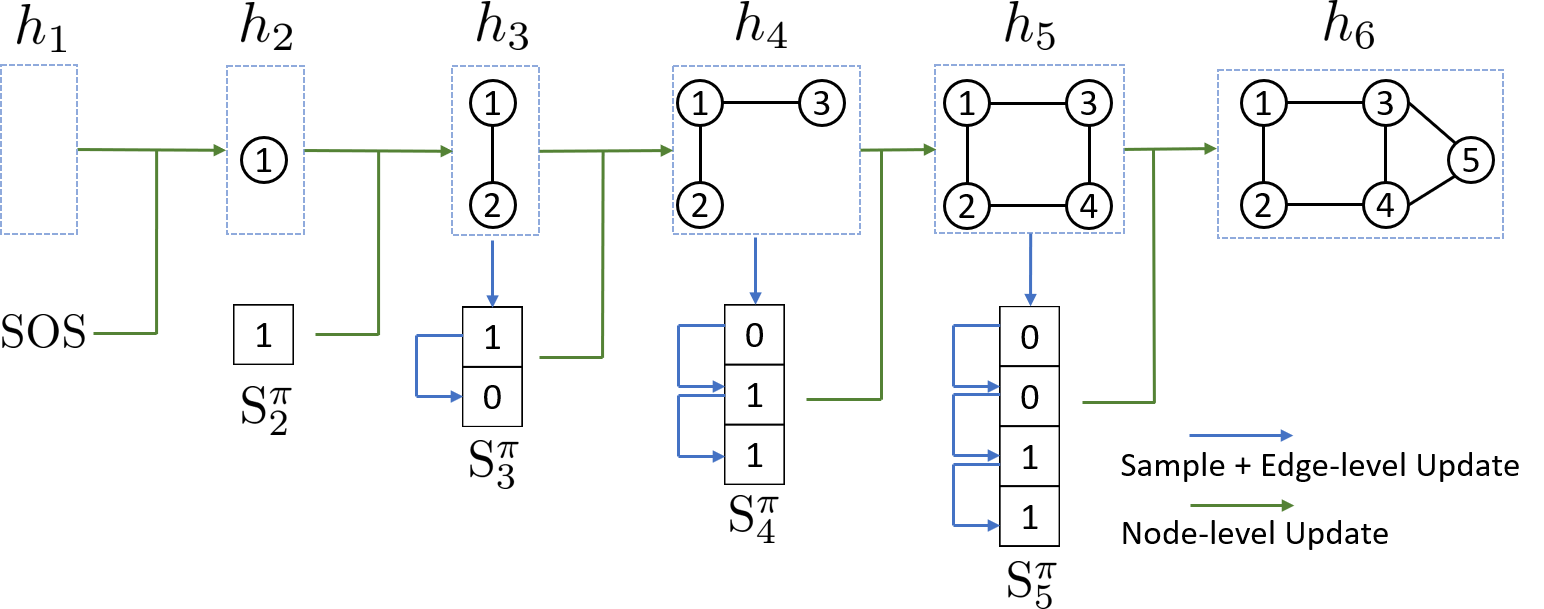}
   \vspace{-10pt}
  \caption{\cut{Graph generation procedure using \graphrnnrnn.} \graphrnnrnn\ at inference time.
     Green arrows denote the graph-level RNN that encodes the ``graph state'' vector $h_i$ in its hidden state, updated by the predicted adjacency vector $S^\pi_{i}$ for node $\pi(v_i)$.
     Blue arrows represent the edge-level RNN, whose hidden state is initialized by the graph-level RNN, that is used to predict the adjacency vector $S^\pi_{i}$ for node $\pi(v_i)$.
   \cut{Green arrows denotes update of hidden state vector of the underlying RNN for $S^\pi$. Blue arrows denote sampling from the predicted probability of $S^\pi_{i, j}=1$, and state vector update given the sample. Neural network weights are shared for the same type of updates}}
  \label{fig:arch}
  \vspace{-10pt}
\end{figure}

The key idea of our approach is to represent graphs under different node orderings as sequences, and then to build an autoregressive generative model on these sequences. \
As we will show, this approach does not suffer from the drawbacks common to other general approaches (\textit{c.f.}, Section \ref{sec:possible}), allowing us to model graphs of varying size with complex edge dependencies, and we introduce a BFS node ordering scheme to drastically reduce the complexity of learning over all possible node sequences (Section \ref{sec:bfs}). 
In this autoregressive framework, the model complexity is greatly reduced by weight sharing with recurrent neural networks (RNNs). 
Figure~\ref{fig:arch} illustrates our \name\ approach, where the main idea is that we decompose graph generation into a process that generates a sequence of nodes (via a {\em graph-level RNN}), and another process that then generates a sequence of edges for each newly added node (via an {\em edge-level RNN}).

\subsubsection{Modeling graphs as sequences}
We first define a mapping $f_S$ from graphs to sequences, where 
for a graph $G\sim p(G)$ with $n$ nodes under node ordering $\pi$, we have
\begin{equation}
  \label{eq:seq_def}
  S^\pi=f_S(G,\pi)=(S^\pi_1,...,S^\pi_{n}),
\end{equation}
where each element $S^\pi_i\in \{0,1\}^{i-1}, i\in \{1,...,n\}$ is an adjacency vector representing the edges between node $\pi(v_{i})$ and the previous nodes $\pi(v_{j}), j\in \{1,...,i-1\}$ already in the graph:\footnote{We prohibit self-loops and  $S^\pi_1$ is defined as an empty vector.}
\begin{equation}
S^\pi_i = (A^\pi_{1,i},...,A^\pi_{i-1,i})^T, \forall i\in \{2,...,n\}.
\end{equation}
For undirected graphs, $S^\pi$ determines a unique graph $G$, and we write the mapping as $f_G(\cdot)$ where $f_G(S^{\pi})=G$. 

Thus, instead of learning $p(G)$, whose sample space cannot be easily characterized, 
we sample the auxiliary $\pi$ to get the observations of $S^\pi$ and learn $p(S^\pi)$, which can be modeled autoregressively due to the sequential nature of $S^\pi$.
At inference time, we can sample $G$ without explicitly computing $p(G)$ by sampling $S^\pi$, which
maps to $G$ via $f_G$.

Given the above definitions, we can write $p(G)$ as the marginal distribution of the joint distribution $p(G,S^\pi)$:
\begin{equation}\label{eq:graphlike}
  p(G) = \sum_{S^\pi}{ p(S^\pi) \ \mathds{1}[f_G(S^\pi)=G]} ,
\end{equation}
where $p(S^\pi)$ is the distribution that we want to learn using a generative model.
Due to the sequential nature of $S^\pi$, we further decompose $p(S^\pi)$ as the product of conditional distributions over the elements: 
\begin{equation}
  p(S^\pi) = \prod_{i=1}^{n+1}{p(S^\pi_i|S^\pi_1,...,S^\pi_{i-1})}
\end{equation}
where we set $S^\pi_{n+1}$ as the end of sequence token $\texttt{EOS}$, to represent sequences with variable lengths. We simplify $p(S^\pi_i|S^\pi_1,...,S^\pi_{i-1})$ as $p(S^\pi_i|S^\pi_{<i})$ in further discussions.
\cut{
\will{Personally, I find this paragraph more confusing than illuminating; it is also odd because we talk about training on all possible sequences but then in Section 2.4 contradict this. 
I would prefer we drop this and just be more explicit in the next section about how we actually train via maximum likelihood. Rather than having this abstract description of the implicit maximum likelihood approach here.}
Note that $p(S^\pi)$ can be trained by maximum likelihood given the observed edge sequences $\mathbb{S}$ of graphs $\mathbb{G}=\{G_1,...,G_s\}$.
More precisely, given a graph $G \in \mathbb{G}$ and assuming all permutations $\pi$ to be equally likely ($p(\pi)=\frac{1}{n!}$), we can then generate a set of observations $\mathbb{S_G}=\{S^{\pi_1}_1,...,S^{\pi_1}_n, .... S^{\pi_k}_{1},...,S^{\pi_k}_{n}\}$ as a set of sequences of edges of nodes in the observed graph $G$ under all $k=|\Pi|$ permutations.
Then given a set of observed edge sequences $\mathbb{S}=\{S_G\}$ of graphs $\mathbb{G}$, we can train the model.
\jure{Note, above we are missing an index: $S$ go over all permutations, over all nodes of each graph and over all graphs -- so $S$ needs to be index by 3 elements (graph, node, permutation).}
}
\begin{algorithm}[t]
   \caption{GraphRNN inference algorithm}
   \label{alg:graphrnn}
\begin{algorithmic}
   \STATE {\bfseries Input:} RNN-based transition module $f_{trans}$, output module $f_{out}$, probability distribution $\mathcal{P}_{\theta_i}$ parameterized by $\theta_i$, start token $\texttt{SOS}$, end token $\texttt{EOS}$, empty graph state $h'$
   \STATE {\bfseries Output:} Graph sequence $S^{\pi}$
   \STATE $S^{\pi}_1=\texttt{SOS}$, $h_1 =  h'$, $i=1$ 
   \REPEAT
   \STATE $i=i+1$
   \STATE $h_i = f_{\mathrm{trans}}( h_{i-1},S^{\pi}_{i-1})$ \COMMENT{update graph state}
   \STATE $\theta_i = f_{\mathrm{out}}(h_i)$
   \STATE $S^{\pi}_i \sim \mathcal{P}_{\theta_i}$ \COMMENT{sample node $i$'s edge connections}
   \UNTIL{$S^{\pi}_i$ is $\texttt{EOS}$}
   \STATE {\bfseries Return} $S^{\pi}=(S^{\pi}_1,...,S^{\pi}_i)$
\end{algorithmic}
\end{algorithm}
\subsubsection{The GraphRNN framework}
So far we have transformed the modeling of $p(G)$ to modeling $p(S^{\pi})$, which we further decomposed into the product of conditional probabilities $p(S^\pi_i|S^\pi_{<i})$. Note that $p(S^\pi_i|S^\pi_{<i})$ is highly complex as it has to capture how node $\pi(v_i)$ links to previous nodes based on how previous nodes are interconnected among each other.
Here we propose to parameterize $p(S^\pi_i|S^\pi_{<i})$ using expressive neural networks to model the complex distribution. To achieve scalable modeling, we let the neural networks share weights across all time steps $i$.
In particular, we use an RNN that consists of a \textit{state-transition} function and an \textit{output} function:
\begin{align}\label{eq:rnnframework}
  h_i &= f_{\mathrm{trans}}({h}_{i-1}, S^\pi_{i-1}),\\
  \theta_{i} &= f_{\mathrm{out}}({h}_i),
\end{align}
where $ h_i \in \R^{d}$ is a vector that encodes the state of the graph generated so far, $S^\pi_{i-1}$ is the adjacency vector for the most recently generated node $i-1$, and $\theta_i$ specifies the distribution of next node's adjacency vector (\ie, $S^{\pi}_i \sim \mathcal{P}_{\theta_i}$).
In general, $f_{\mathrm{trans}}$ and $f_{\mathrm{out}}$ can be arbitrary neural networks, and $\mathcal{P}_{\theta_i}$ can be an arbitrary distribution over binary vectors. 
This general framework is summarized in Algorithm~\ref{alg:graphrnn}.

Note that the proposed problem formulation is fully general; we discuss and present some specific variants with implementation details in the next section.
Note also that RNNs require fixed-size input vectors, while we previously defined $S^\pi_i$ as having varying dimensions depending on $i$; we describe an efficient and flexible scheme to address this issue in
Section \ref{sec:bfs}.


\cut{In our formulation, $\bm h_i$ encodes the necessary information about the currently generated graph and $f_{out}$ does not directly output $S^\pi_i$. Rather, it outputs a vector $\theta_i$ which parametrizes $p(S^\pi_i|S^\pi_{<i})$, and then $S^\pi_i$ is sampled according to the distribution $p(S^\pi_i|S^\pi_{<i})$. 
}

\subsubsection{GraphRNN variants}

Different variants of the GraphRNN model correspond to different assumptions about $p(S^\pi_i|S^\pi_{<i})$. Recall that each dimension of $S^\pi_i$ is a binary value that models existence of an edge between the new node $\pi(v_{i})$ and a previous node $\pi(v_{j}),j\in \{1,...,i-1\}$.
We propose two variants of GraphRNN, both of which implement the transition function $f_{\mathrm{trans}}$ (\ie, the graph-level RNN) as a Gated Recurrent Unit (GRU)~\cite{chung2014empirical} but differ in the implementation of $f_{\mathrm{out}}$ (\ie, the edge-level model).
Both variants are trained using stochastic gradient descent with a maximum likelihood loss over $S^\pi$ --- \ie, we optimize the parameters of the neural networks to optimize $\prod{p_{model}(S^\pi)}$ over all observed graph sequences.

 \cut{
 \jiaxuan{I guess readers can figure out what teacher forcing is, so maybe no need to expand}
 we use a ``teacher forcing'' training scheme where we always update the GRU state using the ground truth adjacency vectors $S^\pi_i$ during training, rather than training on the prediction of the model. 
 }

\cut{
\xhdr{Dependency}
The connections between a new node and previous nodes are dependent (e.g. if trying to model triadic closure property), thus we would expect the random variables in $S^\pi_i$ are correlated.

\xhdr{Multi-modality}
Given a partially generated graph, there are multiple patterns that a new node may connects to previous nodes(e.g. choosing one of the triads to close), making $p(S^\pi_i|S^\pi_{<i})$ multi-modal.}

\xhdr{Multivariate Bernoulli}
First we present a simple baseline variant of our \name approach, which we term \graphrnnmlp\ (``S'' for ``simplified''). 
In this variant, we model $p(S^\pi_i|S^\pi_{<i})$ as a multivariate Bernoulli distribution, parameterized by the $\theta_i\in \mathbb{R}^{i-1}$ vector that is output by $f_{\mathrm{out}}$. 
In particular, we implement $f_{\mathrm{out}}$ as single layer multi-layer perceptron (MLP) with
sigmoid activation function, that shares weights across all time steps. 
The output of $f_{\mathrm{out}}$ is a vector $\theta_i$, whose element $\theta_{i}[j]$ can be interpreted as a probability of edge $(i,j)$. 
We then sample edges in $S^\pi_i$ {\em independently} according to a multivariate Bernoulli distribution parametrized by $\theta_i$.
\cut{
Note that this simple formulation cannot model complex edge dependencies as it assumes each edge in $S^\pi_i$ to be independent. Furthermore, it also assumes that $p(S^\pi_i|S^\pi_{<i})$ has a single mode (since its parameter $\theta_i$ is deterministic given $\bm h_i$). We refer to this formulation as \graphrnnmlp\ (``S'' for ``simple''). \jure{not sure what we mean by ``single mode''. We never defined or mentioned the concept of a ``mode'' before. Not sure why this is relevant.}}


\xhdr{Dependent Bernoulli sequence}
To fully capture complex edge dependencies, in the full \name\ model we further decompose $p(S^\pi_i|S^\pi_{<i})$ into a product of conditionals,
\begin{equation}
  \label{eq:pS_cond_rnn}
  p(S^\pi_i|S^\pi_{<i}) = \prod_{j=1}^{i-1} p(S^\pi_{i,j} | S^\pi_{i,<j}, S^\pi_{<i}),
\end{equation}
where $S^\pi_{i, j}$ denotes a binary scalar that is $1$ if node $\pi(v_{i+1})$ is connected to node $\pi(v_{j})$ (under ordering $\pi$). 
In this variant, each distribution in the product is approximated by an another RNN. 
Conceptually, we have a hierarchical RNN, where the first (\ie, the graph-level) RNN generates the nodes and maintains the state of the graph, while the second (\ie, the edge-level) RNN generates the edges of a given node (as illustrated in Figure~\ref{fig:arch}). 
In our implementation, the edge-level RNN is a GRU model, where the hidden state is initialized via the graph-level hidden state $h_i$ and where the output at each step is mapped by a MLP to a scalar indicating the probability of having an edge. $S^\pi_{i, j}$ is sampled from this distribution specified by the $j$th output of the $i$th edge-level RNN, and is fed into the $j+1$th input of the same RNN. All edge-level RNNs share the same parameters.

\subsubsection{Tractability via breadth-first search}\label{sec:bfs}

A crucial insight in our approach is that rather than learning to generate graphs under any possible node permutation, we learn to generate graphs using breadth-first-search (BFS) node orderings, without a loss of generality. 
Formally, we modify Equation (\ref{eq:seq_def}) to
\begin{equation}
  \label{eq:seq_def_bfs}
  S^\pi=f_S(G,\textsc{BFS}(G,\pi)),
\end{equation}
where $\textsc{BFS}(\cdot)$ denotes the deterministic BFS function.
In particular, this BFS function takes a random permutation $\pi$ as input, picks $\pi(v_1)$ as the starting node and appends the neighbors of a node into the BFS queue in the order defined by $\pi$.
Note that the BFS function is many-to-one, \ie, multiple permutations can map to the same ordering after applying the BFS function. 

Using BFS to specify the node ordering during generation has two essential benefits.
The first is that we only need to train on all possible BFS orderings, rather than all possible node permutations, \ie, multiple node permutations map to the same BFS ordering, providing a reduction in the overall number of sequences we need to consider.\footnote{In the worst case (\eg, star graphs), the number of BFS orderings is $n!$, but we observe substantial reductions on many real-world graphs.} 
The second is that the BFS ordering makes learning easier by reducing the number of edge predictions we need to make in the edge-level RNN; in particular,  when we are adding a new node under a BFS ordering, the only possible edges for this new node are those connecting to nodes that are in the ``frontier'' of the BFS (\ie, nodes that are still in the BFS queue)---a notion formalized by Proposition \ref{prop:bfs_edge} (proof in the Appendix):
\begin{proposition}
\label{prop:bfs_edge}
  Suppose $v_1, \ldots, v_n$ is a BFS ordering of $n$ nodes in graph $G$, and $(v_i, v_{j-1}) \in E$ but $(v_i, v_j) \not \in E$ for some $i < j \le n$,  then $(v_{i'}, v_{j'}) \not \in E$, $\forall 1 \le i' \le i$ and $j \le j' < n$.
\end{proposition}

Importantly, this insight allows us to redefine the variable size $S^\pi_i$ vector as a fixed $M$-dimensional vector, representing the connectivity between node $\pi(v_i)$ and nodes in the current BFS queue with maximum size $M$:
\begin{equation}
S^\pi_i = (A^\pi_{\max(1,i-M),i},...,A^\pi_{i-1,i})^T, i\in \{2, ..., n\}.
\end{equation}
As a consequence of Proposition \ref{prop:bfs_edge}, we can bound $M$ as follows:
\begin{corollary}
With a BFS ordering the maximum number of entries that \name model needs to predict for $S^\pi_i$, $\forall 1 \le i \le n$ is 
$O\left(\max_{d=1}^{\mathrm{diam}(G)} \left|\left\{v_i | \mathrm{dist}(v_i, v_1) = d\right\} \right| \right)$,
where $\mathrm{dist}$ denotes the shortest-path-distance between vertices. 
\end{corollary}
The overall time complexity of \name\ is thus $O(Mn)$.
In practice, we estimate an empirical upper bound for $M$ (see the Appendix for details). 




\cut{
The simple definition of the graph sequence $S^\pi$ in Equation (\ref{eq:seq_def}) poses two challenges:
First, a given graph can be represented by $\frac{n!}{|\mathrm{Aut}(G)|}$ distinct graph sequences,
 where $|\mathrm{Aut}(G)|$ denotes the order of the automorphism group of graph $G$. Since almost all graphs have no non-trivial automorphisms \cite{erdHos1963asymmetric}, the number of different representations for the same graph is almost always close to $n!$ for large $n$.
Second, if we represent $S^\pi_i$ as fixed-size, $M$-dimensional vectors (necessary to implement the graph-level RNN),  then this dimensionality $M$ could restrict the maximum graph size that can be generated.
}
\cut{
To address these issues, we propose a breath-first search (BFS) node ordering scheme, which drastically reduces the number of possible sequences we need to consider by collapsing multiple node orderings to a unique BFS tree.
The tree-structure induced by BFS also allows us to limit the dimensionality of the $S^\pi_i$ vectors. 
Specifically, we modify Equation (\ref{eq:seq_def}) to
\begin{equation}
  \label{eq:seq_def_bfs}
  S^\pi=f_S(G,\textsc{BFS}(G,\pi)),
\end{equation}
where $\textsc{BFS}(\cdot)$ denotes the deterministic BFS function that always picks $\pi(v_1)$ as the starting node and appends the neighbors of a node into the BFS queue in the order defined by $\pi$.
The BFS function collapses many node permutations to the same BFS node order, reducing the number of possible sequences and the cost of approximating Equation \eqref{eq:graphlike}.

In addition, the BFS orderings reduce the required dimensionality $M$ of $S^\pi_i$, as demonstrated in Proposition \ref{prop:bfs_edge}:
\begin{proposition}
\label{prop:bfs_edge}
  Suppose $v_1, \ldots, v_n$ is a BFS ordering of $n$ nodes in graph $G$, and $(v_i, v_{j-1}) \in E$ but $(v_i, v_j) \not \in E$ for some $i < j \le n$,  then $(v_{i'}, v_{j'}) \not \in E$, $\forall 1 \le i' \le i$ and $j \le j' < n$.
\end{proposition}
The proof\footnote{Full proofs are deferred to the Appendix.} relies on the observation that if a node does not connect to a previous node in the BFS ordering, then all its subsequent nodes do not have an edge to that previous node. 
As a consequence, we also have that:
\begin{corollary}
With a BFS ordering the maximum number of entries that \name model needs to predict for $S^\pi_i$, $\forall 1 \le i \le n$ is 
$O\left(\max_{d=1}^{\mathrm{diam}(G)} \left|\left\{v_i | \mathrm{dist}(v_i, v_1) = d\right\} \right| \right)$,
where $\mathrm{dist}$ denotes the shortest-path-distance between vertices. 
\end{corollary}

Following this argument, when implementing both GraphRNN variants, we modify the definition of $S^\pi_i$ as
\begin{equation}
S^\pi_i = (A^\pi_{\max(1,i-M),i},...,A^\pi_{i-1,i})^T, i\in \{2, ..., n\},
\end{equation}
where $M$ is a constant and we zero-pad all $S^\pi_{i}$ to be a length $M$ vector. In this way, using a fixed-size vector only limits the number of previous nodes that a new node can connect to and not the maximum size of the generated graph.
\name thus has overall time complexity $O(Mn)$.
}

%


\cut{
In general, graphs with higher diameter benefit the most from this technique, but in practice, we found that reducing at least half of the dependencies incur negligible performance degradation. \jure{unclear what we mean by ``reducing at least half of the dependencies''.}
}
\cut{
\subsection{Extension to Graphs with Node and Edge Features}
Our \name model can also be applied to graphs where nodes and edges have feature vectors associated with them.
In this extended setting, under node ordering $\pi$, a graph $G$ is associated with its node feature matrix $X^\pi\in\mathbb{R}^{n\times m}$ and edge feature matrix $F^\pi\in\mathbb{R}^{n\times k}$, where $m$ and $k$ are the feature dimensions for node and edge respectively.
Then we can extend the definition of $S^\pi$ to include feature vectors of corresponding nodes as well as edges $S^\pi_i = (X^\pi_i, F^\pi_i)$. We can then model $X^\pi_i$ and $F^\pi_i$ respectively, and instantiate them using MLP, encoder+decoder or RNN.
Note also that directed graphs can be viewed as an undirected graphs with two edge types, which is a special case under the above extension. 
}
\section{\name\ Model Capacity}
\label{sec:case_study}
In this section we analyze the representational capacity of \graphrnnrnn, illustrating how it is able to capture complex edge dependencies. 
In particular, we discuss two very different cases on how \graphrnnrnn\ can learn to generate graphs with a \textit{global community structure} as well as graphs with a very \textit{regular geometric structure}.
For simplicity, we assume that $h_i$ (the hidden state of the graph-level RNN) can exactly encode $S^\pi_{<i}$, and that the edge-level RNN can encode $S^\pi_{i,<j}$.
That is, we assume that our RNNs can maintain memory of the decisions they make and elucidate the models capacity in this ideal case.
We similarly rely on the universal approximation theorem of neural networks \cite{hornik1991approximation}.

\xhdr{Graphs with community structure}
\graphrnnrnn\ can model structures that are specified by a given probabilistic model. This is because the posterior of a new edge probability can be expressed as a function of the outcomes of previous nodes.
For instance, suppose that the training set contains graphs generated from the following distribution $p_{com}(G)$: half of the nodes are in community $A$, and half of the nodes are in community $B$ (in expectation), and nodes are connected with probability $p_s$ within each community and probability $p_d$ between communities.
Given such a model, we have the following key (inductive) observation:
\begin{observation}\label{obs:com}
Assume there exists a parameter setting for \name\ such that it can generate $S^\pi_{<i}$ and $S^\pi_{i,<j}$ according to the distribution over $S^\pi$  implied by  $p_{com}(G)$, then there also exists a parameter setting for \graphrnnrnn\ such that it can output $p(S^\pi_{i, j} | S^\pi_{i, <j}, S^\pi_{<i})$ according to $p_{com}(G)$. 
\end{observation}
This observation follows from three facts:
First, we know that $p(S^\pi_{i, j} | S^\pi_{i, <j}, S^\pi_{<i})$ can be expressed as a function of $p_s$, $p_d$, and $p(\pi(v_j) \in A), p(\pi(v_j) \in B) \: \forall 1 \leq j \le i$ (which holds by $p_{com}$'s definition).
Second, by our earlier assumptions on the RNN memory, $S^\pi_{<i}$ can be encoded into the initial state of the edge-level RNN, and the edge-level RNN can also encode the outcomes of $S^\pi_{i,<j}$.
Third, we know that $p(\pi(v_i) \in A)$ is computable from $S^\pi_{<i}$ and $S^\pi_{i,1}$ (by Bayes' rule and $p_{com}$'s definition, with an analogous result for $p(\pi(v_i) \in B)$). 
Finally, \name\ can handle the  base case of the induction in Observation \ref{obs:com}, \ie, $S_{i,1}$, simply by sampling according to $0.5p_s + 0.5p_d$ at the first step of the edge-level RNN (\ie, 0.5 probability $i$ is in same community as node $\pi(v_1)$). 

\cut{
Next, using the state-transition function of the edge-level RNN, the model can condition on the sampled value of $S^\pi_{i,1}$, and this information, combined with $S^\pi_{<i}$ (which is assumed to be encoded in $h_i$) can be used to compute $p(v_i \in A)$, as well as $p(v_i \in B)$.
Finally, given knowledge of $p(\pi(v_j) \in A), p(\pi(v_j) \in B), \forall 1 \le j $, the model can compute $p(S^\pi_{i,2})$ as
\begin{equation}
p(S^\pi_{i,2}) = 
\end{equation}
}

\cut{
To generate such a graph, \graphrnnmlp\ model makes use of the hidden state that only encodes $S^\pi_{<i}$. Thus, it can learn that the probability of an edge between the next node $\pi(i)$ and any existing node in community $A$ is
\begin{equation}
  \label{eq:prob_next_node_mlp}
  p_a p\left(\pi(v_i) \in A\right) + p_b p(\pi(v_i) \in B).
\end{equation}
}
\cut{
Conditioned on this outcome, and whether node $v_1$ and $v_2$ are in the same community (encoded in the state vector $\bm h_i$), the model can decide if node $v_2$ and $v_1$ are in the same community. 
It then assigns probability $p_a$ or $p_b$, thus preserving the community structure of the network. 
}

\xhdr{Graphs with regular structure}
\graphrnnrnn\ can also naturally learn to generate regular structures, due to its ability to learn functions that only activate for $S^\pi_{i, j}$ where $v_j$ has specific degree.
For example, suppose that the training set consists of ladder graphs~\cite{noy2004recursively}.
To generate a ladder graph, the edge-level RNN must handle three key cases: if $\sum_{k=1}^{j}S^\pi_{i,j} = 0$, then the new node should only connect to the degree $1$ node or else any degree $2$ node; if $\sum_{k=1}^{j}S^\pi_{i,j} = 1$, then the new node should only connect to the degree $2$ node that is exactly two hops away; and finally, if $\sum_{k=1}^{j}S^\pi_{i,j} = 2$ then the new node should make no further connections. 
And note that all of the statistics needed above are computable from $S^\pi_{<i}$ and $S^\pi_{i,<j}$. 
The appendix contains visual illustrations and further discussions on this example. 

\cut{
\rex{i think i do need a figure to explain this clearly. let me do this in appendix first. Here I'll just say how these types of graphs crucially depends on neurons activating for nodes with the right degree, and explain more in appendix} \jure{keeping this in appendix for now sounds great.}
}

\cut{
\xhdr{Discussion: Learning deep generative model of graphs}
Their parameters are dependent only if they do enough GCN rounds.
Their samples are not dependent.
multi-modality.
their formulation is not scalable. \rex{not needed due to shortage of space?}
}


\section{Experiments}
\label{sec:experiments}

\begin{figure*}[t]
    \includegraphics[width=\textwidth]{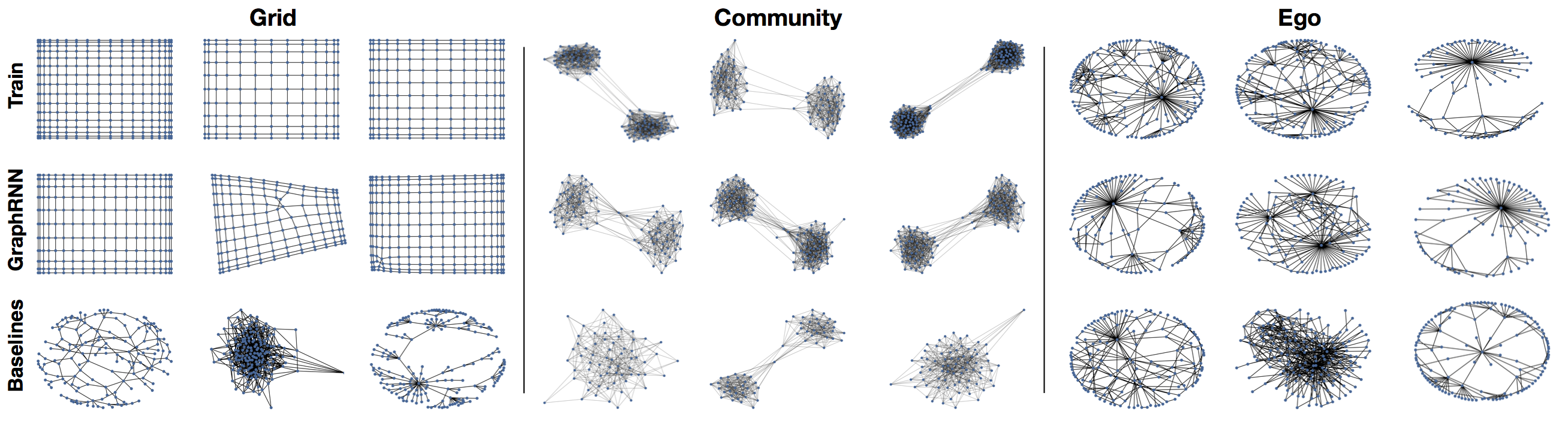}
    \caption{Visualization of graphs from grid dataset (Left group), community dataset (Middle group) and Ego dataset (Right group). Within each group, graphs from training set (First row), graphs generated by \graphrnnrnn (Second row) and graphs generated by Kronecker, MMSB and B-A baselines respectively (Third row) are shown. Different visualization layouts are used for different datasets.}
    \label{fig:graphs_vis}
    \vspace{-10pt}
\end{figure*}

We compare \name\ to state-of-the-art baselines, demonstrating its robustness and ability to generate high-quality graphs in diverse settings. 

\subsection{Datasets}
We perform experiments on both synthetic and real datasets, with drastically varying sizes and characteristics. 
The sizes of graphs vary from $|V|=10$ to $|V|=2025$.

\xhdr{Community} 500 two-community graphs with $60\leq|V|\leq160$. Each community is generated by the \er \ model (E-R) \cite{erdos1959random} with $n=|V|/2$ nodes and $p=0.3$. We then add $0.05|V|$ inter-community edges with uniform probability. 

\xhdr{Grid} 100 standard 2D grid graphs with $100\leq|V|\leq400$. We also run our models on 100 standard 2D grid graphs with $1296\leq|V|\leq2025$, and achieve comparable results.

\xhdr{B-A} 500 graphs with $100\leq|V|\leq200$ that are generated using the \ba model. During generation, each node is connected to 4 existing nodes. 

\xhdr{Protein} 918 protein graphs \cite{dobson2003distinguishing} with $100\leq|V|\leq500$. Each protein is 
represented by a graph, where nodes are amino acids and two nodes are connected if they are less than $6$ Angstroms apart.

\xhdr{Ego}  757 3-hop ego networks extracted from the Citeseer network \cite{sen2008collective} with $50\leq|V|\leq399$. Nodes represent documents and edges represent citation relationships. 

\subsection{Experimental Setup}

We compare the performance of our model against various traditional generative models for graphs, as well as some recent deep graph generative models.

\xhdr{Traditional baselines}
Following \citealt{li2018learning} we compare against the \er\ model (E-R) \cite{erdos1959random} and the \ba (B-A) model \cite{albert2002statistical}.
In addition, we compare against popular generative models that include learnable parameters: Kronecker graph models \cite{leskovec2010kronecker} and mixed-membership stochastic block models (MMSB) \cite{airoldi2008mixed}.

\xhdr{Deep learning baselines} We compare against the recent methods of \citealt{simonovsky2018graphvae} (GraphVAE) and \citealt{li2018learning} (DeepGMG). 
We provide reference implementations for these methods (which do not currently have associated public code), and we adapt GraphVAE to our problem setting by using one-hot indicator vectors as node features for the graph convolutional network encoder.\footnote{We also attempted using degree and clustering coefficients as features for nodes, but did not achieve better performance.}

\xhdr{Experiment settings} We use $80\%$ of the graphs in each dataset for training and test on the rest. 
We set the hyperparameters for baseline methods based on recommendations made in their respective papers. 
The hyperparameter settings for \name\ were fixed after development tests on data that was not used in follow-up evaluations (further details in the Appendix).
Note that all the traditional methods are only designed to learn from a single graph, therefore we train a separate model for each training graph in order to compare with these methods.
In addition, both deep learning baselines suffer from  aforementioned scalability issues, so we only compare to these baselines on a small version of the community dataset with $12\leq|V|\leq20$ (Community-small) and 200 ego graphs with $4\leq|V|\leq18$ (Ego-small).

\begin{table*}[t]
\centering
\begin{footnotesize}
\caption{Comparison of \name to traditional graph generative models using MMD. $(\max(|V|), \max(|E|))$ of each dataset is shown.}
\label{tab:mmd_big}
\begin{tabular}{@{}lllllllllllll@{}}
\toprule
               & \multicolumn{3}{c}{Community (160,1945)} & \multicolumn{3}{c}{Ego (399,1071)}  & \multicolumn{3}{c}{Grid (361,684)} & \multicolumn{3}{c}{Protein (500,1575)} \\ \cmidrule(lr){2-4}\cmidrule(lr){5-7}\cmidrule(lr){8-10}\cmidrule(lr){11-13}
               & Deg.   & Clus.   & Orbit   & Deg.   & Clus.  & Orbit & Deg.   & Clus.   & Orbit & Deg.   & Clus.   & Orbit \\ \midrule
E-R          &0.021 &1.243 &0.049 &0.508 &1.288 &0.232 &1.011    &0.018    &0.900    &0.145 &1.779 &1.135 \\
B-A          &0.268 &0.322 &0.047 &0.275 &0.973 &0.095 &1.860    &0        &0.720    &1.401 &1.706 &0.920 \\
Kronecker  &0.259 &1.685 & 0.069 & 0.108 &0.975 &0.052 &1.074    &0.008    &0.080    &0.084 &0.441 &0.288 \\
MMSB       &0.166 &1.59 &0.054 & 0.304 &0.245 &0.048 &1.881    &0.131    &1.239    &0.236 &0.495 &0.775 \\ \midrule
\graphrnnmlp     &0.055 &0.016 &0.041 &0.090 & \textbf{0.006} &0.043 & 0.029    &$10^{-5}$&0.011    &0.057 & \textbf{0.102} & \textbf{0.037} \\
\graphrnnrnn & \textbf{0.014} & \textbf{0.002} & \textbf{0.039} & \textbf{0.077} & 0.316 & \textbf{0.030} & $\mathbf{10^{-5}}$ & \textbf{0}        & $\mathbf{10^{-4}}$ & \textbf{0.034} &0.935 &0.217\\ \bottomrule
\end{tabular}
\end{footnotesize}
\end{table*}

\begin{table*}[t]
\centering
\begin{footnotesize}
\caption{\name compared to state-of-the-art deep graph generative models on small graph datasets using MMD and negative log-likelihood (NLL). $(\max(|V|), \max(|E|))$ of each dataset is shown. (DeepVAE and GraphVAE cannot scale to the graphs in Table \ref{tab:mmd_big}.)}
\label{tab:mmd_small}
\begin{tabular}{@{}lllllllllll@{}}
\toprule
               & \multicolumn{5}{c}{Community-small (20,83)} & \multicolumn{5}{c}{Ego-small (18,69)} \\ \cmidrule(lr){2-4}\cmidrule(lr){5-6}\cmidrule(lr){7-9}\cmidrule(lr){10-11}
               & Degree & Clustering  & Orbit & Train NLL& Test NLL   & Degree   & Clustering   & Orbit &Train NLL & Test NLL  \\ \midrule
GraphVAE    &0.35 &0.98 &0.54 &13.55  &25.48  & 0.13   & 0.17      &   0.05    &   12.45    &   14.28      \\
DeepGMG     &0.22 &0.95 &0.40 &106.09 &112.19 &0.04  &0.10  & 0.02 &21.17  &22.40\\
\graphrnnmlp    &\textbf{0.02} &0.15 &\textbf{0.01} &31.24  &35.94  &0.002 &\textbf{0.05}  &\textbf{0.0009} &8.51   &9.88    \\
\graphrnnrnn &0.03 &\textbf{0.03} &\textbf{0.01} &28.95  &35.10  &\textbf{0.0003}&\textbf{0.05}  &\textbf{0.0009} &9.05   &10.61    \\ \bottomrule
\end{tabular}
\end{footnotesize}
\end{table*}

\subsection{Evaluating the Generated Graphs}

Evaluating the sample quality of generative models is a challenging task in general \cite{Theis2016a}, and in our case, this evaluation requires a comparison between two sets of graphs (the generated graphs and the test sets).
Whereas previous works relied on qualitative visual inspection \cite{simonovsky2018graphvae} or simple comparisons of average statistics between the two sets \cite{leskovec2010kronecker}, we propose novel evaluation metrics that compare all moments of their empirical distributions. 
\cut{
Previous works proposed to evaluate the generated graphs by visual inspection \cite{simonovsky2018graphvae} or by comparing the average statistics over the two sets \cite{leskovec2010kronecker}. However, visual examination is solely qualitative while average statistics discard the higher moments of the data. Ideally, we would like to quantitatively evaluate the similarity between two sets by comparing all moments of their empirical distributions.}

Our proposed metrics are based on Maximum Mean Discrepancy (MMD) measures.
Suppose that a unit ball in a reproducing kernel Hilbert space (RKHS) $\mathcal{H}$ is used as its function class $\mathcal{F}$,
and $k$ is the associated kernel, the squared MMD between two sets of samples from distributions $p$ and $q$ can be derived as \cite{gretton2012kernel}
\begin{equation}
\label{eq:mmd}
\begin{split}
\mmd^2(p||q)& =\mathbb{E}_{x,y\sim p}[k(x,y)]+\mathbb{E}_{x,y\sim q}[k(x,y)]\\
& -2\mathbb{E}_{x\sim p,y\sim q}[k(x, y)].
\end{split}
\end{equation}
Proper distance metrics over graphs are in general computationally intractable \cite{lin1994hardness}.
Thus, we compute MMD using a set of graph statistics $\mathbb{M}=\{M_1,...,M_k\}$, where each $M_i(G)$ is a univariate distribution over $\mathbb{R}$, such as the degree distribution or clustering coefficient distribution. 
We then use the first Wasserstein distance as an efficient distance metric between two distributions $p$ and $q$:
\begin{equation}
\label{eq:emd}
W(p, q) = \inf_{\gamma \in \Pi(p, q)} \mathbb{E}_{(x, y) \sim \gamma} [||x - y||],
\end{equation}
where $\Pi(p, q)$ is the set of all distributions whose marginals are $p$ and $q$ respectively, and $\gamma$ is a valid transport plan.
To capture high-order moments, we use the following kernel, whose Taylor expansion is a linear
combination of all moments (proof in the Appendix):
\begin{proposition}
  \label{ref:mmd_emd}
  The kernel function defined by $k_W(p, q) = \exp{\frac{W(p, q)}{2\sigma^2}}$ induces a unique RKHS.
\end{proposition}\cut{
As proven in \citealt{kolouri2016sliced}, this Wasserstein distance based kernel is a positive definite (p.d.) kernel. 
By properties that linear combinations, 
product and limit (if exists) of p.d. kernels are p.d. kernels, $k_W(p, q)$ is also a p.d. kernel.\footnote{This can be seen by expressing the kernel function using Taylor expansion.}
By the Moore-Aronszajn theorem, a symmetric p.d. kernel induces a unique RKHS. Therefore Equation
\eqref{eq:mmd} holds if we set $k$ to be $k_W$.}

In experiments, we show this derived MMD score for degree and clustering coefficient distributions, as well as average orbit counts statistics, \emph{i.e.}, the number of occurrences of all orbits with 4 nodes (to capture higher-level motifs) \cite{hovcevar2014combinatorial}. We use the RBF kernel to compute distances between count vectors.

\subsection{Generating High Quality Graphs}
Our experiments demonstrate that \name can generate graphs that match the characteristics of the ground truth graphs in a variety of metrics.

\xhdr{Graph visualization}
Figure \ref{fig:graphs_vis} visualizes the graphs generated by \name\ and various baselines, showing that \name\ can capture the structure of datasets with vastly differing characteristics---being able to effectively learn regular structures like grids as well as
more natural structures like ego networks. Specifically, we found that grids generated by \name\
do not appear in the training set, \ie, it learns to generalize to unseen grid widths/heights.

\cut{
\xhdr{Graph visualization}
We first visualize some examples of training graphs and graphs generated by \name and baseline methods on 3 datasets in Figure \ref{fig:graphs_vis}. We select the generated graphs so that none of the generated graphs are in the training set, showing the generalization ability of our models.
Without hand-crafted design, \name can effectively learn from datasets with distinct characteristics, such as grid structure, community structure and high degree hub, and generate similar graphs accordingly. 
Among the tasks, generating grids is especially hard, as the model should precisely capture the regular structure of grids without making any errors.
\name can perform extremely well in grid dataset, owing to the conditional information captured by the model as discussed in section 3.
In comparison, MMSB only performs well on the community dataset, Kronecker model only performs reasonable on the ego network, while B-A model performs badly in all datasets.
}
	
\begin{figure}[t]
\vspace{-10pt}
    \centering
    \begin{subfigure}[b]{0.23\textwidth}
        \includegraphics[width=\textwidth]{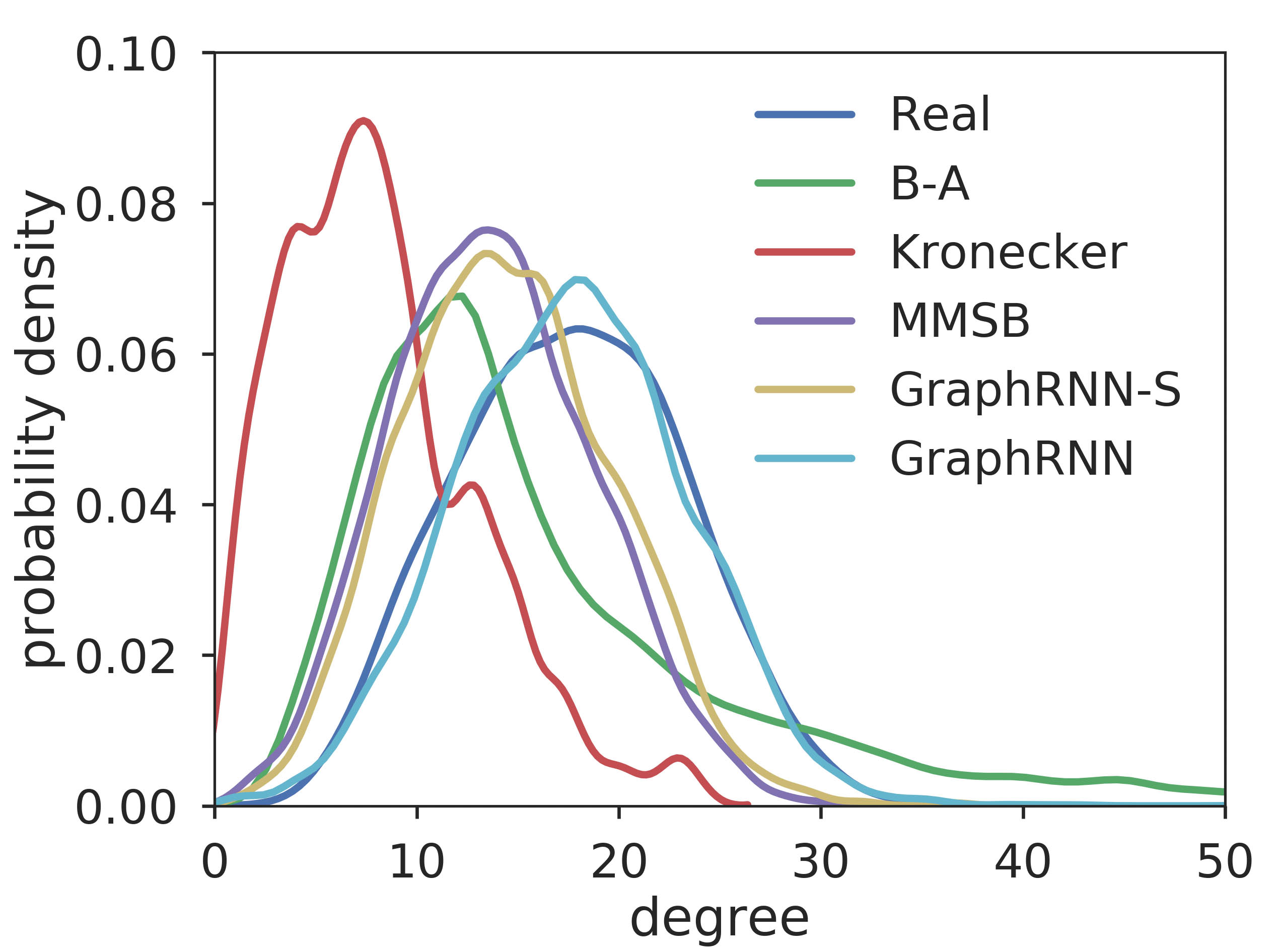}
        \label{fig:robustness_degree}
    \end{subfigure}
    \begin{subfigure}[b]{0.23\textwidth}
        \includegraphics[width=\textwidth]{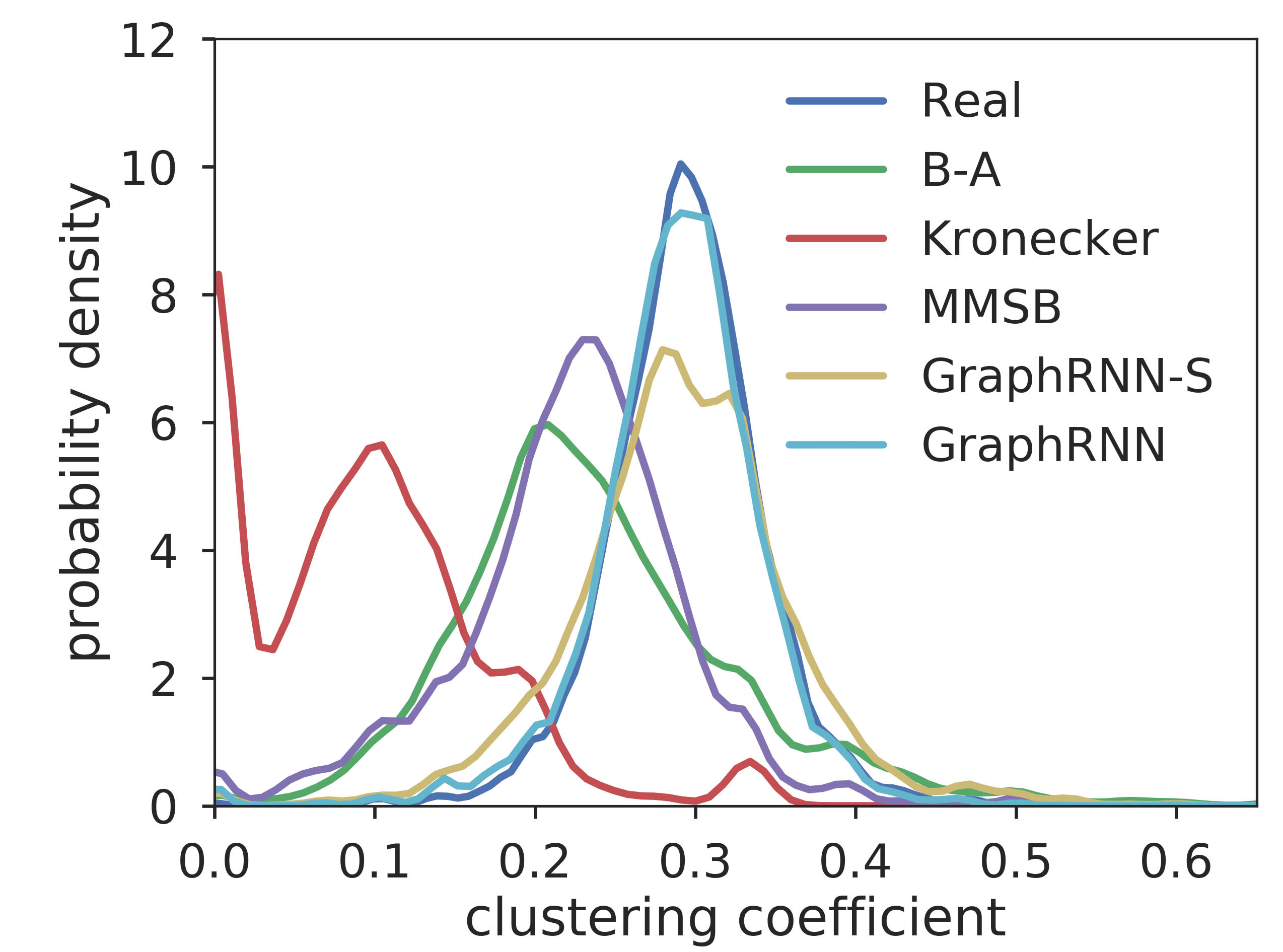}
        \label{fig:robustness_clustering}
    \end{subfigure}
    \vspace{-20pt}
    \caption{Average degree (Left) and clustering coefficient (Right) distributions of graphs from test set and graphs generated by GraphRNN and baseline models.}
    \label{fig:average}
\end{figure}

\xhdr{Evaluation with graph statistics}
We use three graph statistics---based on degrees, clustering coefficients and orbit counts---to further quantitatively evaluate the generated graphs.
Figure~\ref{fig:average} shows the average graph statistics in the test vs.\@ generated graphs, which demonstrates that even from hundreds of graphs with diverse sizes, \graphrnnrnn\ can still learn to capture the underlying graph statistics very well, with the generated average statistics closely matching the overall test set distribution. 

Tables \ref{tab:mmd_big} and \ref{tab:mmd_small} summarize MMD evaluations on the full datasets and small versions, respectively. 
Note that we train all the models with a fixed number of steps, and report the test set performance at the step with the lowest training error.\footnote{Using the training set or a validation set to evaluate MMD gave analogous results, so we used the train set for early stopping.}
GraphRNN variants achieve the best performance on all datasets, with $80\%$ decrease of MMD on average compared with traditional baselines, and $90\%$ decrease of MMD compared with deep learning baselines.
Interestingly, on the protein dataset, our simpler \graphrnnmlp\ model performs very well, which is likely due to the fact that the protein dataset is a nearest neighbor graph over Euclidean space and thus does not involve highly complex edge dependencies. 
Note that even though some baseline models perform well on specific datasets (\eg, MMSB on the community dataset), they fail to generalize across other types of input graphs. 

\xhdr{Generalization ability}
Table \ref{tab:mmd_small} also shows negative log-likelihoods (NLLs) on the training and test sets. We report the average $p(S^\pi)$ in our model, and report the likelihood in baseline methods as defined in their papers. A model with good generalization ability should have small NLL gap between training and test graphs. We found that our model can generalize well, with $22\%$ smaller average NLL gap.\footnote{The average likelihood is ill-defined for the traditional models.}\cut{, while maintaining high sample qualities, with  \todo{xx} decrease of average MMD.}

\subsection{Robustness}
\begin{figure}[t]
 \vspace{-10pt}
    \centering
    \begin{subfigure}[b]{0.22\textwidth}
        \includegraphics[width=\textwidth]{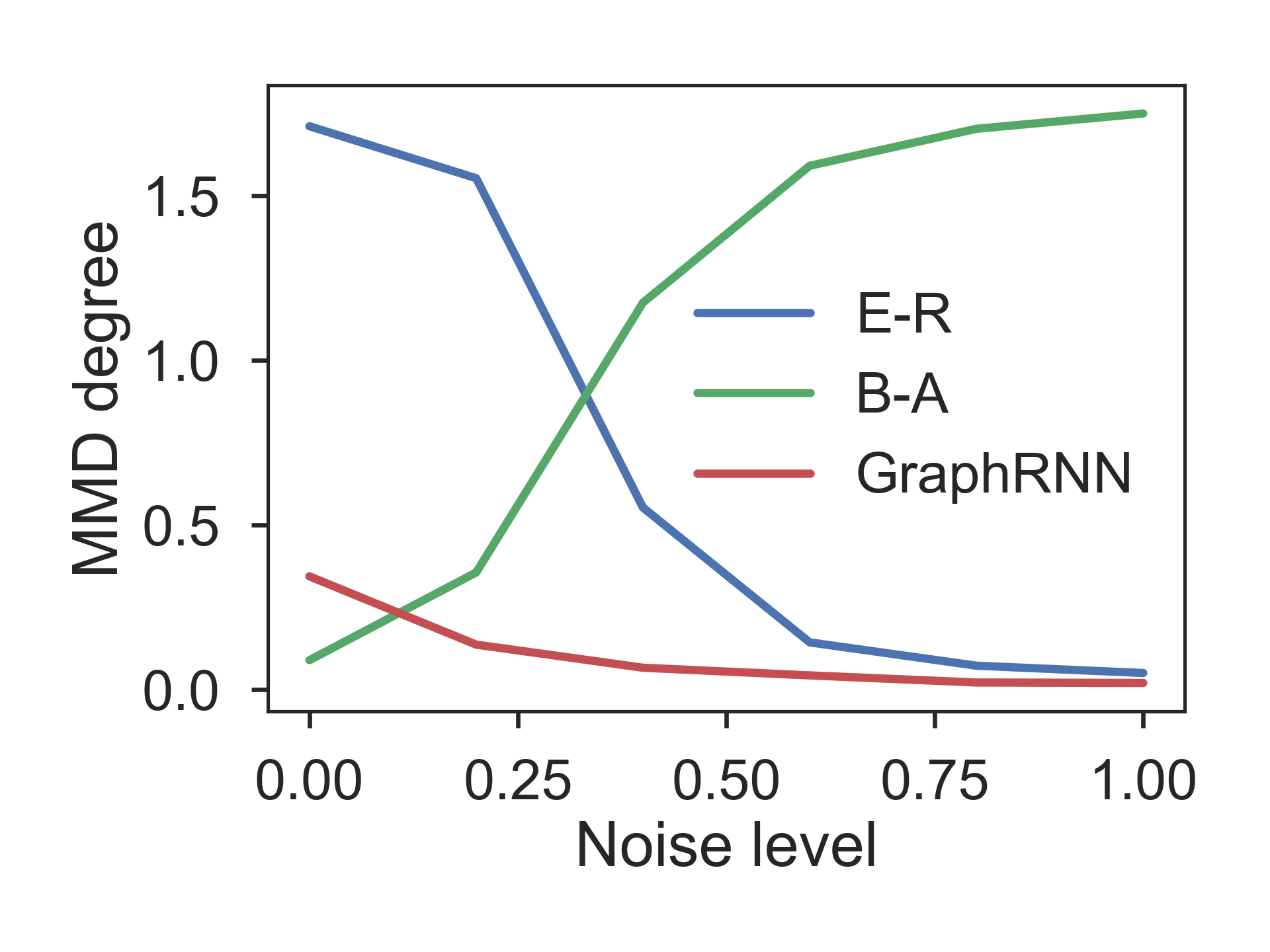}
        \label{fig:robustness_degree}
    \end{subfigure}
    \begin{subfigure}[b]{0.22\textwidth}
        \includegraphics[width=\textwidth]{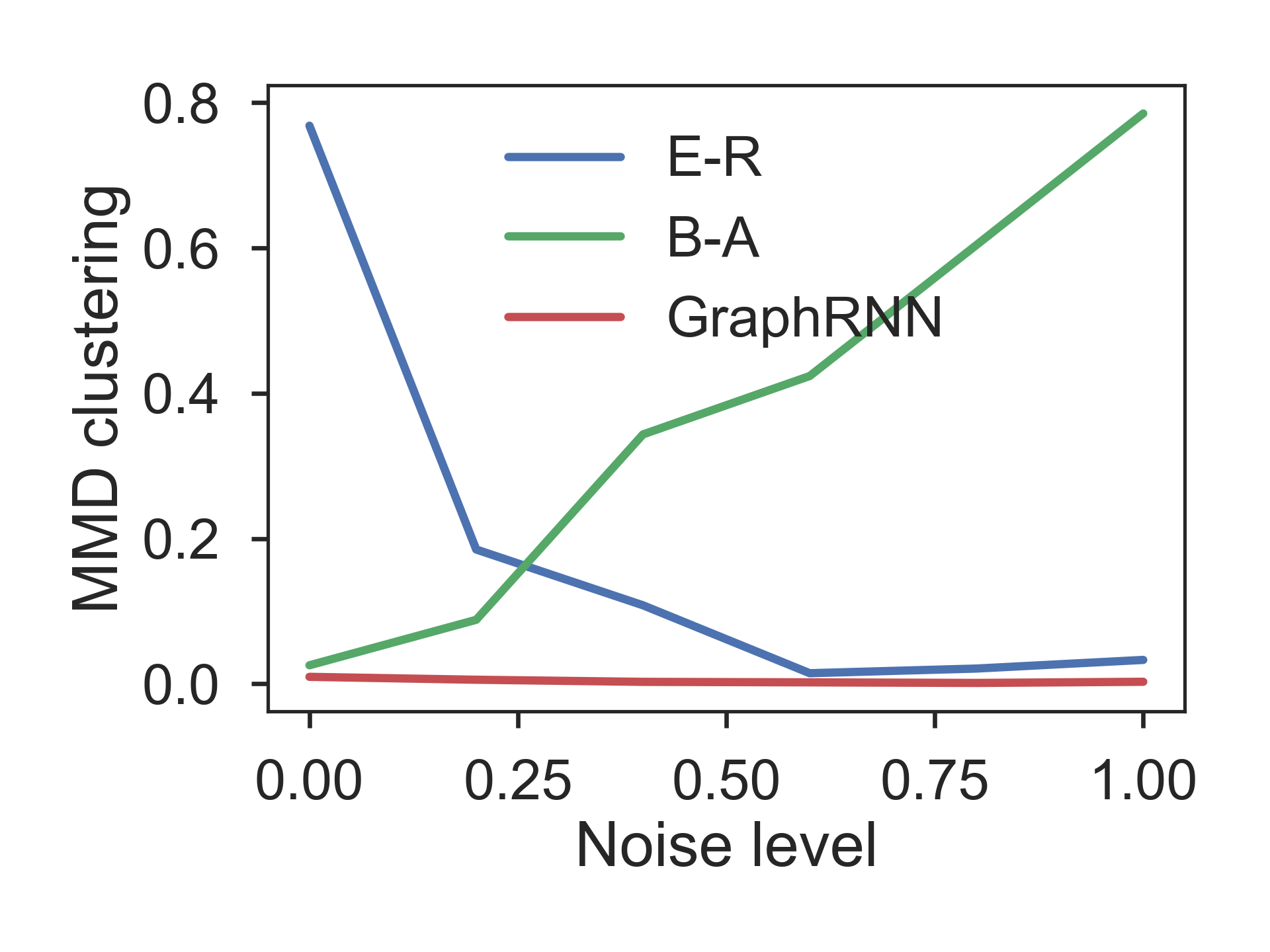}
        \label{fig:robustness_clustering}
    \end{subfigure}
       \vspace{-25pt}
    \caption{MMD performance of different approaches on degree (Left) and clustering coefficient (Right) under different noise level.}
    \label{fig:robustness}
\end{figure}

Finally, we also investigate the robustness of our model by interpolating between \ba (B-A) and \er\ (E-R) graphs. We randomly perturb [$0\%, 20\%,...,100\%$] edges of B-A graphs with $100$ nodes. With $0\%$ edges perturbed, the graphs are E-R graphs; with $100\%$ edges perturbed, the graphs are B-A graphs. 
Figure \ref{fig:robustness} shows the MMD scores for degree and clustering coefficient distributions for the $6$ sets of graphs.
Both B-A and E-R perform well when graphs are generated from their respective distributions, but their performance degrades significantly once noise is introduced. 
In contrast, \name maintains strong performance as we interpolate between these structures, indicating high robustness and versatility.

\section{Further Related Work}


In addition to the deep graph generative approaches and traditional graph generation approaches surveyed previously, our framework also builds off a variety of other methods.

\cut{
\xhdr{Traditional graph generative models}
Many generative models in the past focus on specific properties of a particular class of graphs, and design generation processes that cater to these properties. Preferential attachment models such as \ba graphs model the power law of degree and clustering distributions of many realistic networks \cite{albert2002statistical}. The forest fire model \cite{leskovec2007graph} aims to capture the properties such as heavy-tail degree distributions and shrinking diameters of social networks. However, these models have few parameters and are not flexible enough to model realistic graphs which possess a variety of different properties.
Kronecker graph models \cite{leskovec2010kronecker} utilize the fractal structure of certain graphs to fit degree, clustering and diameter statistics, but again this approach cannot model graphs with different underlying generation mechanisms.
Exponential random graph models (ERGMs) have also proved extremely useful as a statistical model of real-world graphs, where edge probabilities are defined by a learned exponential-family distribution; however, the ERGM framework is designed for a fixed number of nodes and relies heavily on knowledge of domain-specific node attributes \cite{robins2007introduction}. 
\rex{Jure thinks that it can be removed. Just add some of the points into the traditional generative models discussion in intro.}
}

\cut{
\xhdr{Deep graph generative models} 
Recently, a number of deep learning based models for generating graphs have been proposed.
\citealt{kipf2016variational} and \citealt{grovergraphite} both propose graph generation models based on deep variational autoencoders, but these methods are limited to generating graphs of a fixed size and cannot train on multiple graphs without explicit node alignment.
More recently, \citealt{simonovsky2018graphvae} proposed GraphVAE, which
uses an encoder-decoder framework to directly generate adjacency matrix representations of graphs, while \citealt{li2018learning} propose an approach based upon iterative graph convolutions.
These recent methods face significant limitations due to strong conditional independence assumptions that are made and involve prohibitively high computational complexity, being unable to scale to graphs with $>40$ nodes.
}

\xhdr{Molecule and parse-tree generation}
There has been related domain-specific work on generating candidate molecules and parse trees in natural language processing. 
Most previous work on discovering molecule structures make use of a expert-crafted sequence representations of molecular graph structures (SMILES) \cite{olivecrona2017molecular,segler2017generating,gomez2016automatic}.
Most recently, SD-VAE \cite{dai2018syntax-directed} introduced a grammar-based approach to generate structured data, including molecules and parse trees. 
In contrast to these works, we consider the fully general graph generation setting without assuming features or special structures of graphs.

\cut{
\xhdr{Exponential random graphs models} Exponential random graph models (ERGMs) are a powerful statistical tool for modeling real-world graphs, where edge probabilities are defined by a learned exponential-family distribution. 
However, unlike our approach the ERGM framework is designed for inference over a fixed number of nodes and relies heavily on knowledge of domain-specific node attributes \cite{robins2007introduction}. 
}

\xhdr{Deep autoregressive models}
Deep autoregressive models decompose joint probability distributions as a product of conditionals, a general idea that has achieved striking successes in the image \cite{oord2016pixel} and audio \cite{oord2016wavenet} domains.\cut{
For example, PixelRNN \cite{oord2016pixel} can generate highly realistic images by decomposing the joint distribution of pixels into conditional probability distributions of individual pixels, while
WaveNet \cite{oord2016wavenet} proposes an autoregressive model for time signals that achieves state-of-the-art performance on audio generation.}
Our approach extends these successes to the domain of generating graphs. 
Note that the DeepGMG algorithm \cite{li2018learning} and the related prior work of \citealt{johnson2016learning} can also be viewed as deep autoregressive models of graphs. 
However, unlike these methods, we focus on providing a scalable (\ie, $O(n^2)$) algorithm that can generate general graphs. 

\cut{
\will{I don't think you need related work for the evaluation part.}
\xhdr{Evaluation of generative models}
}

\section{Conclusion and Future Work}
\label{sec:conclusion}
We proposed GraphRNN, an autoregressive generative model for graph-structured data, along with a comprehensive evaluation suite for the graph generation problem, which we used to show that GraphRNN achieves significantly better performance compared to previous state-of-the-art models, while being scalable and robust to noise.
However, significant challenges remain in this space, such as scaling to even larger graphs and developing models that are capable of doing efficient conditional graph generation.

\section*{Acknowledgements}
\label{sec:ack}
The authors thank Ethan Steinberg, Bowen Liu, Marinka Zitnik and Srijan Kumar for their helpful discussions and comments on the paper. This research has been supported in part by DARPA SIMPLEX, ARO MURI, Stanford Data Science
Initiative, Huawei, JD, and Chan Zuckerberg Biohub. W.L.H. was also supported by the SAP Stanford
Graduate Fellowship and an NSERC PGS-D grant.

\bibliography{graph}
\bibliographystyle{icml2018}

\appendix
\section{Appendix}

\subsection{Implementation Details of \name}
In this section we detail parameter setting, data preparation and training strategies for \name. 

We use two sets of model parameters for GraphRNN. 
One larger model is used to train and test on the larger datasets that are used to compare with traditional methods.
One smaller model is used to train and test on datasets with nodes up to $20$. This model is only used to compare with the two most recent preliminary deep generative models for graphs proposed in \cite{li2018learning,simonovsky2018graphvae}.

For \graphrnnrnn, the graph-level RNN uses $4$ layers of GRU cells, with $128$ dimensional hidden state for the larger model, and $64$ dimensional hidden state for the smaller model in each layer. 
The edge-level RNN uses $4$ layers of GRU cells, with $16$ dimensional hidden state for both the larger model and the smaller model. To output the adjacency vector prediction, the edge-level RNN first maps the highest layer of the $16$ dimensional hidden state to a $8$ dimensional vector through a MLP with ReLU activation, then another MLP maps the vector to a scalar with sigmoid activation.
The edge-level RNN is initialized by the output of the graph-level RNN at the start of generating $S^\pi_i$, $\forall 1 \le i \le n$.
Specifically, the highest layer hidden state of the graph-level RNN is used to initialize the lowest layer of edge-level RNN, with a liner layer to match the dimensionality.
During training time, teacher forcing is used for both graph-level and edge-level RNNs, i.e., we use the groud truth rather than the model's own prediction during training.
At inference time, the model uses its own preditions at each time steps to generate a graph.

For the simple version \graphrnnmlp, a two-layer MLP with ReLU and sigmoid activations respectively is used to generate $S^\pi_i$, with $64$ dimensional hidden state for the larger model, and $32$ dimensional hidden state for the smaller model.
In practice, we find that the performance of the model is relatively stable with respect to these hyperparameters.

We generate the graph sequences used for training the model following the procedure in Section 2.3.4. Specifically, we first randomly sample a graph from the training set, then randomly permute the node ordering of the graph. We then do the deterministic BFS discussed in Section 2.3.4 over the graph with random node ordering, resulting a graph with BFS node ordering. An exception is in the robustness section, where we use the node ordering that generates B-A graphs to get graph sequences, in order to see if GraphRNN can capture the underlying preferential attachment properties of B-A graphs.

\begin{figure*}[t]
    \centering
     \includegraphics[width=0.75\textwidth]{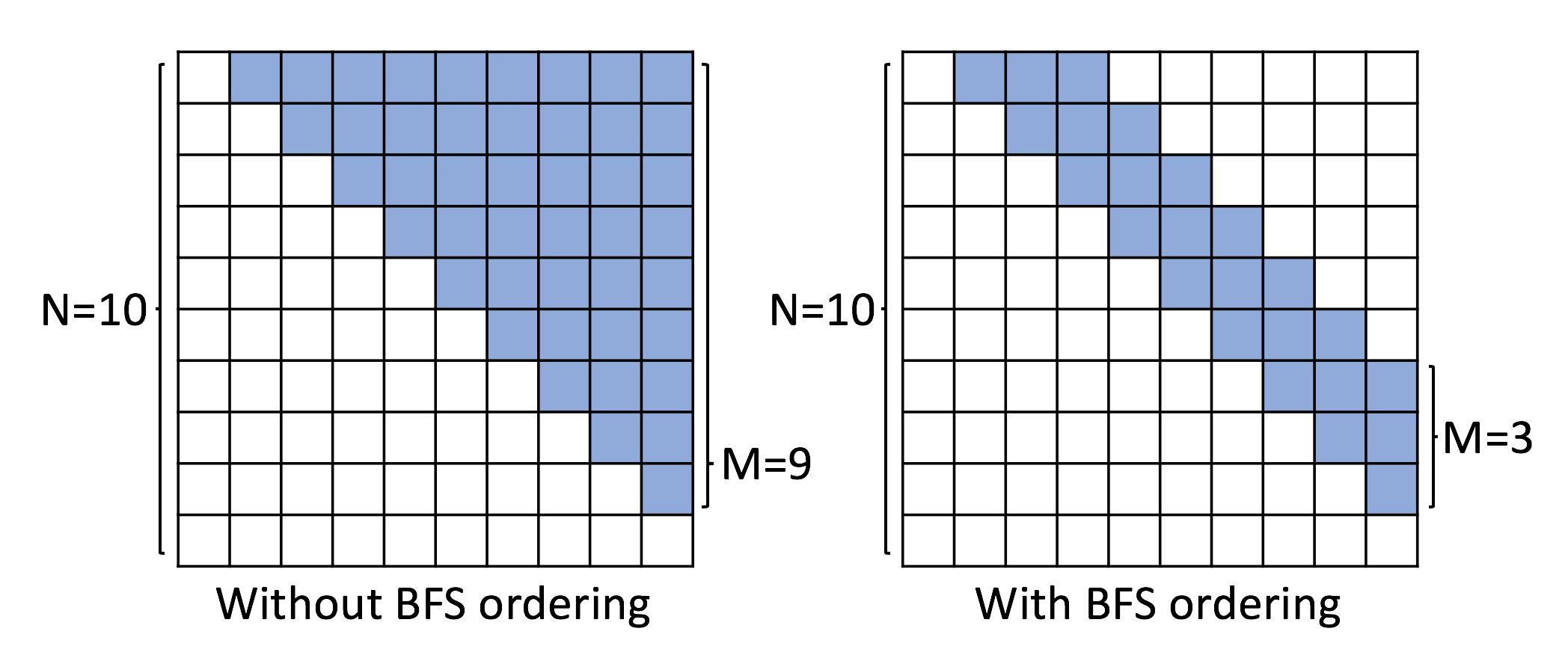}
      \caption{Illustrative example of reducing the maximum dimension $M$ of $S^\pi_i$ through the BFS node ordering. Here we show the adjacency matrix of a graph with $N=10$ nodes. Without the BFS node ordering (Left), we have to set $M=N-1$ to encode all the necessary connection information (shown in dark square). With the BFS node ordering, we could set $M$ to be a constant smaller than $N$ (we show $M=3$ in the figure).}
    \label{fig:bfs}
\end{figure*}

With the proposed BFS node ordering, we can reduce the maximum dimension $M$ of $S^\pi_i$, illustrated in Figure \ref{fig:bfs}. To set the maximum dimension $M$ of $S^\pi_i$, we use the following empirical procedure. We randomly ran $100000$ times the above data pre-processing procedure to get graph with BFS node orderings. We remove the all consecutive zeros in all resulting $S^\pi_i$, to find the empirical distribution of the dimensionality of $S^\pi_i$. We set $M$ to be roughly the $99.9$ percentile, to account for the majority dimensionality of $S^\pi_i$. In principle, we find that graphs with regular structures tend to have smaller $M$, while random graphs or community graphs tend to have larger $M$. Specifically, for community dataset, we set $M=100$; for grid dataset, we set $M=40$; for B-A dataset, we set $M=130$; for protein dataset, we set $M=230$; for ego dataset, we set $M=250$; for all small graph datasets, we set $M=20$.

The Adam Optimizer is used for minibatch training. Each minibatch contains $32$ graph sequences. 
We train the model for $96000$ batchs in all experiments.
We set the learning rate to be $0.001$, which is decayed by $0.3$ at step $12800$ and $32000$ in all experiments.

\subsection{Running Time of \name}

Training is performed on only $1$ Titan X GPU. For the protein dataset that consists of about $1000$ graphs, each containing about $500$ nodes, training converges at around $64000$ iterations. The runtime is around $12$ to $24$ hours. This also includes pre-processing, batching and BFS, which are currently implemented using CPU without multi-threading. The less expressive \graphrnnmlp\ variant is about twice faster.
At inference time, for the above dataset, generating a graph using the trained model only takes about $1$ second.

\subsection{More Details on \name's Expressiveness}

\begin{figure}[h]
     \centering
     \includegraphics[width=0.47\textwidth]{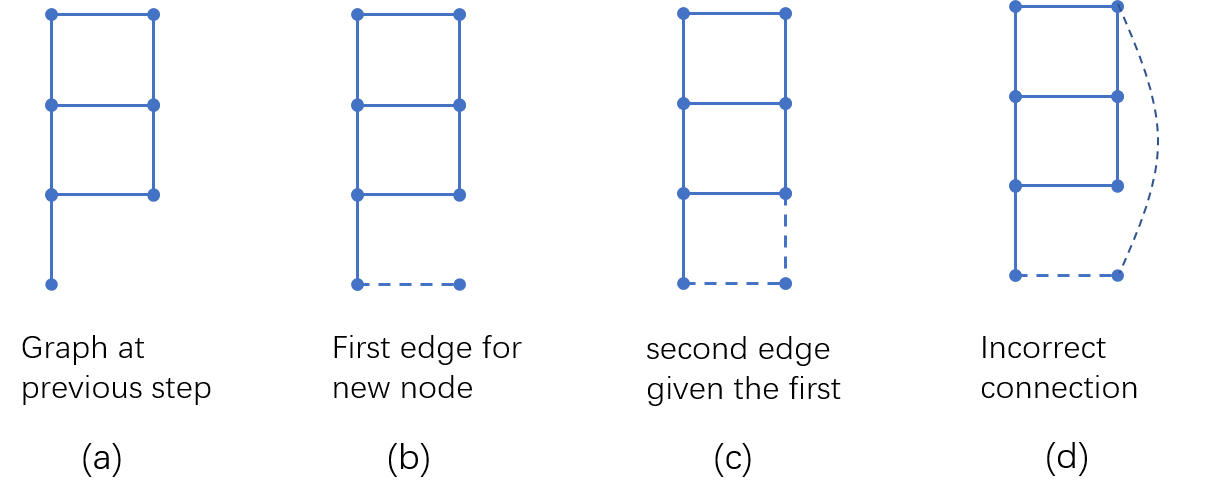}
\vspace{-0.25cm}     
     \caption{Illustration that generation of ladder networks relies on dependencies modeled by \name.}
     \label{fig:capacity_ladder}
 \end{figure}

We illustrate the intuition underlying the good performance of \graphrnnrnn on graphs with regular
structures, such as grid and ladder networks.
Figure \ref{fig:capacity_ladder} (a) shows the generation process of a ladder graph at 
an intermediate step.
At this time step, the ground truth data (under BFS node ordering) specifies that the new node added to the graph should make an edge to the node with degree $1$. Note that node degree is a function of $S^\pi_{<i}$, thus could be approximated by a neural network.

Once the first edge has been generated, the new node should make an edge with another node of degree $2$. However, there are multiple ways to do so, but only one of them gives a valid grid structure,
\emph{i.e.} one that forms a $4$-cycle with the new edge.
\name\ crucially relies on the edge-level RNN and the knowledge of the previously added edge, in order to distinguish between the correct and incorrect connections in Figure \ref{fig:capacity_ladder} (c) and (d).

\vspace{-0.15cm}

\subsection{Code Overview}

In the code repository, \texttt{main.py} consists of the main training pipeline, which loads datasets and performs training and inference. It also consists of the \texttt{Args} class, which stores the hyper-parameter settings of the model.
\texttt{model.py} consists of the RNN, MLP and loss function modules that are use to build \name. 
\texttt{data.py} contains the minibatch sampler, which samples a random BFS ordering of a batch of randomly selected graphs.
\texttt{evaluate.py} contains the code for evaluating the generated graphs using the MMD metric introduced in Sec. 4.3.

Baselines including the \er\ model, \ba\ model, MMSB, and rge very recent deep generative models (GraphVAE, DeepGMG) are also implemented in the \texttt{baselines} folders. 
We adopt the C++ Kronecker graph model implementation in the SNAP package \footnote{The SNAP package is available at \url{http://snap.stanford.edu/snap/index.html}.}. 

\subsection{Proofs}

\subsubsection{Proof of Proposition 1}
We use the following observation:
\begin{observation}
\label{ob:bfs_order}
By definition of BFS, if $i < k$, then the children of $v_i$ in the BFS ordering come before the children of $v_k$ that do not connect to $v_{i'}$, $\forall 1 \le i' \le i$. 
\end{observation}

By definition of BFS, all neighbors of a node $v_i$ include the parent of $v_i$ in the BFS tree, all children of $v_i$ which have consecutive indices, and some children of $v_{i'}$ which connect to both $v_{i'}$ and $v_i$, for some $1 \le i' \le i$.
Hence if $(v_i, v_{j-1}) \in E$ but $(v_i, v_j) \not \in E$, $v_{j-1}$ is the last children of $v_i$ in the BFS ordering. Hence $(v_{i}, v_{j'}) \not \in E$, $\forall j \le j' \le n$.

For all $i' \in [i]$, supposed that $(v_{i'}, v_{j'-1}) \in E$ but $(v_{i'}, v_{j'}) \not \in E$. By Observation \ref{ob:bfs_order}, $j' < j$. 
By conclusion in the previous paragraph, $(v_{i'}, v_{j''}) \not \in E$, $\forall j' \le j'' \le n$.
Specifically, $(v_{i'}, v_{j''}) \not \in E$, $\forall j \le j'' \le n$. 
This is true for all $i' \in [i]$. 
Hence we prove that 
$(v_{i'}, v_{j'}) \not \in E$, $\forall 1 \le i' \le i$ and $j \le j' < n$.

\subsubsection{Proof of Proposition 2}

As proven in \citealt{kolouri2016sliced}, this Wasserstein distance based kernel is a positive definite (p.d.) kernel. 
By properties that linear combinations, 
product and limit (if exists) of p.d. kernels are p.d. kernels, $k_W(p, q)$ is also a p.d. kernel.\footnote{This can be seen by expressing the kernel function using Taylor expansion.}
By the Moore-Aronszajn theorem, a symmetric p.d. kernel induces a unique RKHS. Therefore Equation
(9) holds if we set $k$ to be $k_W$.

\subsection{Extension to Graphs with Node and Edge Features}
Our \name model can also be applied to graphs where nodes and edges have feature vectors associated with them.
In this extended setting, under node ordering $\pi$, a graph $G$ is associated with its node feature matrix $X^\pi\in\mathbb{R}^{n\times m}$ and edge feature matrix $F^\pi\in\mathbb{R}^{n\times k}$, where $m$ and $k$ are the feature dimensions for node and edge respectively.
In this case, we can extend the definition of $S^\pi$ to include feature vectors of corresponding nodes as well as edges $S^\pi_i = (X^\pi_i, F^\pi_i)$. We can adapt the $f_{out}$ module, by using a MLP to generate $X^\pi_i$ and an edge-level RNN to genearte $F^\pi_i$ respectively.
Note also that directed graphs can be viewed as an undirected graphs with two edge types, which is a special case under the above extension. 

\subsection{Extension to Graphs with Four Communities}
\begin{figure}[t]
    \centering
     \includegraphics[width=0.95\columnwidth]{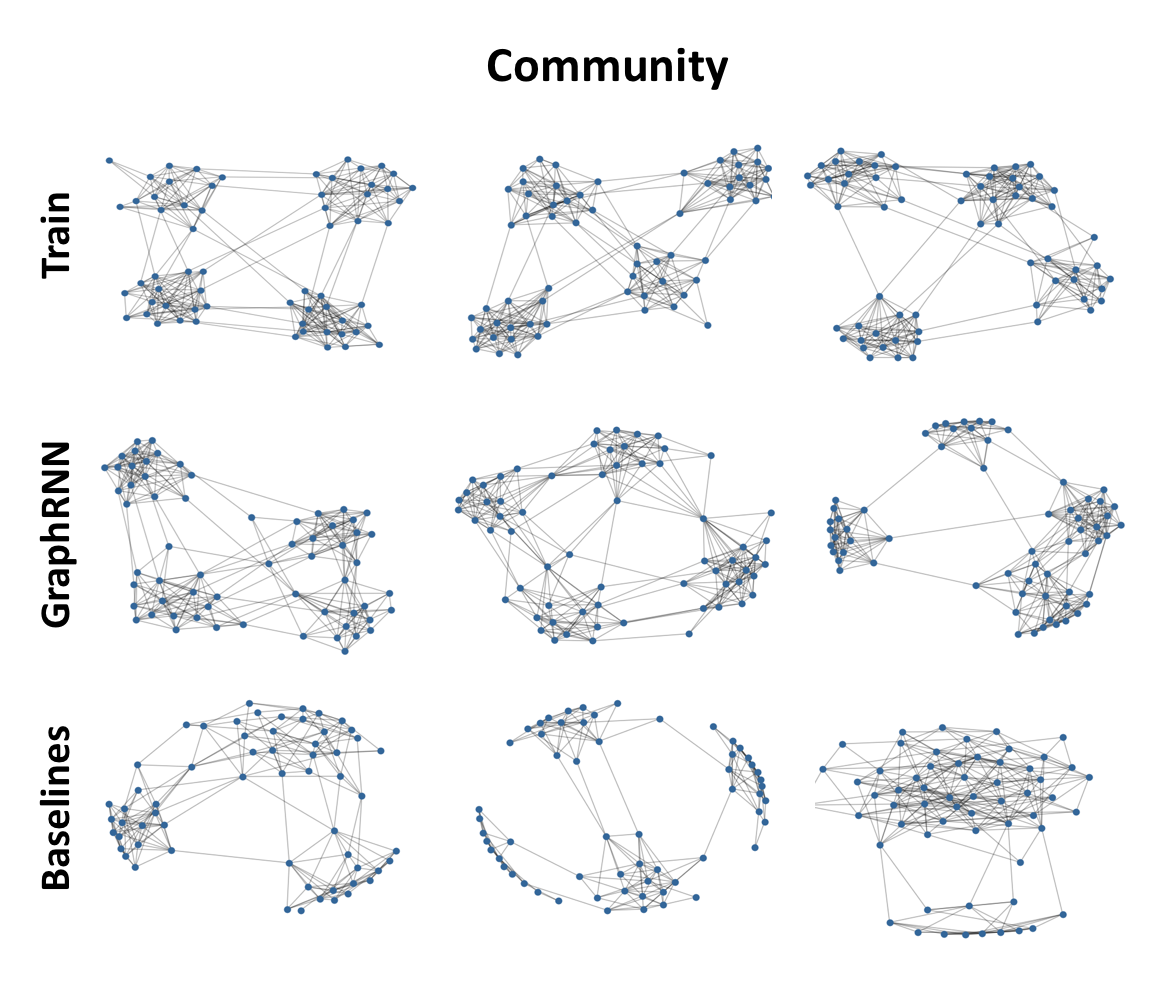}
      \caption{Visualization of graph dataset with four communities. Graphs from training set (First row), graphs generated by \graphrnnrnn (Second row) and graphs generated by Kronecker, MMSB and B-A baselines respectively (Third row) are shown.}
    \label{fig:4community}
\end{figure}

To further show the ability of \name to learn from community graphs, we further conduct experiments on a four-community synthetic graph dataset. Specifically, the data set consists of 500 four community graphs with $48\leq|V|\leq68$. Each community is generated by the \er \ model (E-R) \cite{erdos1959random} with $n \in [|V|/4-2, |V|/4+2]$ nodes and $p=0.7$. We then add $0.01|V|^2$ inter-community edges with uniform probability. FIgure \ref{fig:4community} shows the comparison of visualization of generated graph using \name and other baselines. We observe that in contrast to baselines, \name consistently generate 4-community graphs and each community has similar structure to that in the training set.

\end{document}